%% file: main.tex
\documentclass[11pt,letterpaper]{article}
\usepackage{acl2012}
\usepackage{times}
\usepackage{latexsym}

\usepackage{mathtools}
\usepackage{url}
\usepackage{amsmath}
\usepackage{multirow}
\usepackage[normalem]{ulem}
\usepackage{color}
\usepackage[usenames,dvipsnames]{xcolor}
\usepackage{enumitem}
\usepackage[lined,boxed,commentsnumbered]{algorithm2e}
\DontPrintSemicolon

\usepackage{graphicx}
\usepackage{epstopdf}
\usepackage{subfig}
\usepackage{mdwlist}
\usepackage{enumitem}
\usepackage{amsfonts}
\usepackage{amsmath}
\usepackage{mathtools}
\usepackage{setspace}
\usepackage{stackengine}
\usepackage{array}

\title{Winning on the Merits:\\ The Joint Effects of Content and Style on Debate Outcomes}

\author{Lu Wang$^{1}$ ~~~~ Nick Beauchamp$^{2,3}$ ~~~~ Sarah Shugars$^{3}$  ~~~~ Kechen Qin$^{1}$\\
  $^{1}$College of Computer and Information Science, Northeastern University \\
  $^{2}$Department of Political Science, Northeastern University \\
  $^{3}$Network Science Institute, Northeastern University \\
{\tt luwang@ccs.neu.edu},~{\tt n.beauchamp@northeastern.edu}\\
{\tt \{qin.ke,shugars.s\}@husky.neu.edu}
   \\}

\begin{document}

\maketitle

\input{abstract.tex}

\section{Introduction}
\input{intro.tex}

\section{Data Description}
\label{sec:data}
\input{data.tex}

\section{The Debate Prediction Model}
\label{sec:model}
\input{model.tex}

\section{Experimental Results}
\label{sec:experiment}
\input{prediction.tex}

\section{Discussion}
\label{sec:furtherdiscuss}
\input{discussion.tex}

\section{Related Work}
\label{sec:related}
\input{related.tex}

\section{Conclusion}
\label{sec:conclusion}
\input{conclusion.tex}

\section*{Acknowledgments}
This work was supported in part by National Science Foundation Grant IIS-1566382 and a GPU gift from Nvidia. We thank the TACL reviewers for valuable suggestions on various aspects of this work.

\bibliography{debate,other}
\bibliographystyle{acl2012}

\end{document}

%% file: abstract.tex
\begin{abstract}
\fontsize{10}{11}\selectfont
Debate and deliberation play essential roles in politics and government, but most models presume that debates are won mainly via superior style or agenda control. Ideally, however, debates would be won on the merits, as a function of which side has the stronger arguments.  We propose a predictive model of debate that estimates the effects of linguistic features and the latent persuasive strengths of different topics, as well as the interactions between the two. Using a dataset of 118 Oxford-style debates, our model's combination of content (as latent topics) and style (as linguistic features) allows us to predict audience-adjudicated winners with 74\% accuracy, significantly outperforming linguistic features alone (66\%). Our model finds that winning sides employ stronger arguments, and allows us to identify the linguistic features associated with strong or weak arguments.
\end{abstract}

%% file: intro.tex
What determines the outcome of a debate?  In an ideal setting, a debate is a mechanism for determining which side has the better arguments and for an audience to reevaluate their views in light of what they have learned. This ideal vision of debate and deliberation has taken an increasingly central role in modern theories of democracy \cite{habermas1984theory,cohen1989deliberation,rawls1997idea,mansbridge2003rethinking}. However, empirical evidence has also led to an increasing awareness of the dangers of style and rhetoric in biasing participants towards the most skillful, charismatic, or numerous speakers \cite{noelleNeumann1974,sunstein1999}.  

In light of these concerns, most efforts to predict the persuasive effects of debate have focused on the linguistic features of debate speech~\cite{katzav2008modelling,mochales2011argumentation,feng2011classifying} or on simple measures of topic control~\cite{dryzek2003social,mansbridge2015minimalist,zhangetal:NAACL2016}. In the ideal setting, however, we would wish for the winning side to win based on the strength and merits of their arguments, not based on their skillful deployment of linguistic style. Our model therefore predicts debate outcomes by modeling not just the persuasive effects of directly observable linguistic features, but also by incorporating the inherent, latent strengths of topics and issues specific to each side of a debate.

\begin{figure}[t]
	{\fontsize{9}{10}\selectfont
    \setlength{\tabcolsep}{0.6mm}
    \hspace{-1mm}
    \begin{tabular}{|p{78mm}|}
    \hline
    
    \underline{\textbf{Motion}}: {\it Abolish the Death Penalty} \\
	$\bullet$ Argument 1 (\textsc{pro}): 
	What is the error rate of convicting people that are innocent?
	%Sam Gross at the University of Michigan Law School submitted a paper in the National Academy of Science Proceedings and showed that 
	...when you look at capital convictions, you can demonstrate on innocence grounds a 4.1 percent error rate, 4.1 percent error rate. I mean, would you accept that in flying airplanes? I mean, really. ...
	%Now, there's no doubt innocents have also been executed. 
\\
	$\bullet$ Argument 2 (\textsc{con}):
%There are a handful of cases, though, of people who were really innocent and who were really sentenced to death. But ask yourself where would those people be if they had been sentenced to life without parole instead? In most cases they would still be in prison and they would never get out, because 
%...an inmate who is sentenced to death gets a government paid lawyer for the second review of his case and the third review of his case and that is where this evidence often comes out. Someone sentenced to life in prison, he may have longer time to prove innocence, but he doesn't have the resources, and so in most cases nobody's going to care about his case. 
... The risk of an innocent person dying in prison and never getting out is greater if he's sentenced to life in prison than it is if he's sentenced to death. So the death penalty is an important part of our system. 
\\
	$\bullet$ Argument 3 (\textsc{pro}): 
	%I think one of the sad things about the death penalty is because the punishment is so severe that the limited resources -- as much as Barry is able to do with the Innocence Projects across the country, have to focus on, you know, those people who are sentenced to death.  
	...I think if there were no death penalty, there would be many more resources and much more opportunity to look for and address the question of innocence of people who are serving other sentences.
\\
	
	\hline
    \end{tabular}
    }
\vspace{-3mm}
\caption{\small In this segment from a debate over abolishing the death penalty, argument 1 is identified as having the linguistic features `questions' and `numerical evidence', while arguments 2 and 3 use `logical reasoning.' Our model infers that both pro arguments are intrinsically ``strong,'' while the con argument is ``weak'', although arguments 2 and 3 both use `reasoning' language.}
%Sample arguments on ``innocent people" from debate of ``Abolish the Death Penalty". In argument 1 (a1), pro side brings up the problem of convicting innocent people by using rhetorical questions (in \textcolor{OliveGreen}{\bf green}) and numbers \textcolor{red} (in {\bf red}). Con side uses logical reasoning which is highlighed in \textcolor{blue}{\bf blue} as response. Argument 3 (a3) from pro side also uses logical reasoning to address argument 2 (a2). Our system infers this topic as ``strong" for pro and ``weak" for con. Both a1 and a3 are from pro, but are of different rhetorical traits. Meanwhile, though a2 and a3 both utilize reasoning, a2 is relatively weaker within this context.}
\label{fig:example_intro}
\end{figure}

%The study of debates has been limited on one hand by inadequate measures of actual persuasive effects, and on the other by simple models that investigate either stylistic elements~\cite{katzav2008modelling,greene2009syntactic,mochales2011argumentation,feng2011classifying} or coarse measures of topic control~\cite{dryzek2003social,mansbridge2015minimalist,zhangetal:NAACL2016,card2015media}. 
%

To illustrate this idea, Figure~\ref{fig:example_intro} shows a brief exchange from a debate about the death penalty. Although the arguments from both sides are on the same subtopic (the execution of innocents), they make their points with a variety of stylistic maneuvers, including rhetorical questions, factual numbers, and logical phrasing. Underlying these features is a shared content, the idea of the execution of innocents. Consistent with the work of ~\newcite{baumgartner2008decline}, this subtopic appears to inherently support one side -- the side opposed to the death penalty -- more strongly than the other, \textit{independent of stylistic presentation}.
We hypothesize that within the overall umbrella of a debate, some topics will tend to be inherently more persuasive for one side than the other, such as the execution of innocents for those opposed to the death penalty, or the gory details of a murder for those in favor of it.  Strategically, debaters seek to introduce topics that are strong for them and weaker for their opponent, while also working to craft the most persuasive stylistic delivery they can.  Because these stylistic features themselves vary with the inherent strength of topics, we are able to predict the latent strength of topics even in new debates on entirely different issues, allowing us to predict debate winners with greater accuracy than before.

In this paper, then, we examine \textit{the latent persuasive strength of debate topics, how they interact with linguistic styles, and how both predict debate outcomes}.
%A dataset of 118 Oxford-style debates with recorded audience opinion both before and after the debate is employed for modeling and conducting analysis. 
%to model debates and their outcomes at sufficient scale to understand these complex interactions. 
Although the task here is fundamentally predictive, it is motivated by the following substantive questions: 
How do debates persuade listeners?  By merit of their content and not just their linguistic structure, can we capture the sense in which debates are an exchange of arguments that are strong or weak? How do these more superficial linguistic or stylistic features interact with the latent persuasive effects of topical content? Answering these questions is crucial for modern theories of democratic representation, although we seek only to understand how these features predict audience reactions in the context of a formal debates where substance perhaps has the best chance of overcoming pure style. We discuss in detail the relevance of our work to existing research on framing, agenda setting, debate, persuasion, and argument mining in \S~\ref{sec:related}.

%% NB: SEEMS TOO APOLOGETIC
%\footnote{Instead of aiming to fully solve the problems, our goal is to provide empirical evidence on the interplay between rhetoric and information, and whether our modeling carries predictive power for persuasion.}
%Previous debate persuasion modeling~\cite{boydstun2014realtime} focuses on collecting audience response to debate topics as well as debater's deployed rhetorical frames. However, it is limited by the relative scarcity of debates and does not focus on the interactions between content and rhetoric. Recently, there is a growing interest in detecting arguments and their strength. For instance, pipeline models~\cite{stab2014identifying,persin2015modeling} are developed to identify argument components and assess their strength based on rich linguistic features for student essays. Nevertheless, none of them addresses the persuasiveness of arguments.

%Outside of the debate context, \newcite{higgins2004evaluating} measure the global and local coherence of each sentence of a student essay to assess its relative strength towards supporting the core claim, while pipeline models~\cite{stab2014identifying,persin2015modeling} have been developed to identify argument components and assess each component's strength based on rich linguistic features. However, none of them jointly models the arguments and their strength, nor do they adequately account for the interaction between debaters.

% contribution
We develop here a joint model that simultaneously 1) infers the latent persuasive strength inherent in debate topics and how it differs between opposing sides, and 2) captures the interactive dynamics between topics of different strength and the linguistic structures with which those topics are presented. %Our dataset of 118 Oxford-style debates contains debate transcripts as well as aggregate audience opinion both before and after the debate, thereby allowing us to use the audience response to estimate the latent persuasion of different debate subtopics, the effects of numerous linguistic features, and the interaction between the two.
%
%By directly optimizing for debate outcome, our model also learns to predict argument persuasiveness with regard to audience opinion change. 
%The debate outcome is used as supervision to drive the learning process. 
%A dataset of 118 Oxford-style debates with recorded audience opinion both before and after the debate is employed for modeling and conducting analysis. 
%
% main results
Experimental results on a dataset of 118 Oxford-style debates show that our topic-aware debate prediction model achieves an accuracy of 73.7\% in predicting the winning side. 
% where winning is defined as the side that gains the most support over the course of the debate. 
This result is significantly better than a classifier trained with only linguistic features (66.1\%), or using audience feedback (applause and laughter; 58.5\%), and significantly outperforms previous predictive work using the same data~\cite{zhangetal:NAACL2016} (73\% vs. 65\%). This shows that the inherent persuasive effect of argument content plays a crucial role in affecting the outcome of debates.% and hence in deliberative learning and democracy.

% argument usage: winners tend to use more arguments of topics with strong strength
% shifting: tend to shift to strong topics, and especially the ones favor themselves but not the other side
% features: 
Moreover, we find that winning sides are more likely to have used inherently strong topics (as inferred by the model) than losing sides (59.5\% vs. 54.5\%), a result echoed by human ratings of topics without knowing debate outcomes (44.4\% vs. 30.1\%). Winning sides also tend to shift discussion to topics that are strong for themselves and weak for their opponents. 
Finally, our model is able to identify  linguistic features that are specifically associated with strong or weak topics. For instance, when speakers are using inherently stronger topics, they tend to use more first person plurals and negative emotion, whereas when using weaker arguments, they tend to use more second person pronouns and positive language.  These associations are what allow us to predict topic strength, and hence debate outcomes, out of sample even for debates on entirely new issues.

%% file: data.tex
This study uses transcripts from Intelligence Squared U.S. (IQ2) debates.\footnote{http://intelligencesquaredus.org} %Recorded live in New York City, 
Each debate brings together panels of renowned experts to argue for or against a given issue before a live audience. The debates cover a range of current issues in politics and policy, and are intended to ``restore civility, reasoned analysis, and constructive public discourse.''
Following the Oxford style of debate, each side is given a 7-minute opening statement. The moderator then begins the discussion phase, allowing questions from the audience and panelists, followed by 2-minute closing statements.  

The live audience members record their pre- and post-debate opinions as \textsc{pro}, \textsc{con}, or \textsc{undecided} relative to the resolution under debate. The results are shared only after the debate has concluded. %We collected 118 debates in our Intelligence Squared corpus, all with clear winning side.
According to IQ2, the winning side is the one that gains the most votes after the debate. 118 debate transcripts were collected, with \textsc{pro} winning 60.\footnote{\newcite{zhangetal:NAACL2016} also use IQ2 to study talking points and their predictive power on debate outcome. Our dataset includes theirs plus 11 additional debates (excluding one with result as tie).} 84 debates had two debaters on each side, and the rest had three per side. Each debate contains about 255 speaking turns and 17,513 words on average.\footnote{The dataset can be downloaded from \url{http://www.ccs.neu.edu/home/luwang/}.} 
%with 71.7\% of the words spoken by debaters.

These debates are considerably more structured and moderated, and have more educated speakers and audience members, than one generally finds in public debates. As such, prediction results of our model may vary on other types of debates. 
%As such, while they are an ideal setting to test our model of argument strength, the results here might constitute an upper bound on the predictive accuracy of such models.  On the other hand, 
Meanwhile, since we do not focus on formal logic and reasoning structure, but rather on the intrinsic persuasiveness of different topics, it may be that the results here are more general to all types of argument. Answering this question depends on subsequent work comparing debates of varying degrees of formality.

%These debates are considerably more structured and moderated, and have more educated speakers and audience members, than one generally finds in public debates, let alone more free-form interpersonal arguments.  As such, while they are an ideal setting to test our model of argument strength, the results here might constitute an upper bound on the predictive accuracy of such models.  On the other hand, since we do not focus on formal logic and reasoning structure, but rather on the intrinsic persuasiveness of different topics, it may be that the results here are more general to all types of argument.  Answering this question will depend on subsequent work comparing debates of varying degrees of formality.

%% file: model.tex
We consider debate as a process of argument exchange. %An argument, for the present purposes, is a sequence of contiguous text on a single (sub)topic. 
Arguments have content with inherent (or exogenously determined) persuasive effects as well as a variety of linguistic features shaping that content. %While the effects of these linguistic features have been well-examined already (see \S~\ref{sec:related}), it is also the case that argument content itself has inherent (or exogenously determined) persuasive effects. 
%In a debate, assuming the debaters are familiar with the topics and their inherent effects, the debating process is a persuasive game where debaters choose which topics to deploy based on their strength (constrained by social expectations to stay ``on topic") and then package those arguments in most persuasive language they can (constrained by topic-language interactions).
%
We present here a debate outcome prediction model that combines directly observed linguistic features with latent persuasive effects specific to different topical content. 
%The problem is defined in \S~\ref{sec:problem}. A joint learning framework and features are described in \S~\ref{sec:learning} and \S~\ref{sec:features}. We then present topic strength inference in \S~\ref{sec:inference}, followed by argument extraction (\S~\ref{sec:argextraction}).

\subsection{Problem Statement}
\label{sec:problem}
Assume that the corpus $D$ contains $N$ debates, where $D=\{d_{i}\}_{i=1}^{N}$. Each debate $d_{i}$ comprises a sequence of arguments, denoted as $\mathbf{x_{i}}=\{x_{i,j}\}_{j=1}^{n_{i}}$, where $n_{i}$ is the number of arguments. For the present purposes, an argument is a continuous unit of text on the same topic, and may contain multiple sentences within a given turn (see Figure~\ref{fig:example_intro}). %Argument extraction is described in \S~\ref{sec:argextraction}. 
We use $\mathbf{x_{i}^{p}}$ and $\mathbf{x_{i}^{c}}$ to denote arguments for \textsc{pro} and \textsc{con}. 
The outcome for debate $d_{i}$ is $y_{i}\in \{1, -1\}$, where 1 indicates \textsc{pro} wins and -1 indicates \textsc{con} wins.

%Each debate $d_{i}$ comprises two sets of arguments, \textsc{pro} and \textsc{con}, denoted as $\mathbf{x_{i}}=\{\mathbf{x_{i}^{p}}, \mathbf{x_{i}^{c}}\}$. We have $\mathbf{x_{i}^{p}}=\{x_{i,1}^{p}, x_{i,2}^{p}, \ldots, x_{i,n_{i}^{p}}^{p}\}$ as arguments from \textsc{pro}, and $\mathbf{x_{i}^{c}}=\{x_{i,1}^{c}, x_{i,2}^{c}, \ldots, x_{i,n_{i}^{c}}^{c}\}$ as arguments from \textsc{con}, where $n_{i}^{p}$ and $n_{i}^{c}$ are the number of arguments each side. For the present purposes, an argument is a continuous text unit within a turn and may contain multiple sentences on the same topic (see Figure~\ref{fig:example_intro}). We describe how to extract the arguments in \S~\ref{sec:argextraction}. 

We assume that each debate $d_{i}$ has a topic system, where debaters issue arguments from $K$ topics relevant to the motion ($K$ varies for different debates). Each topic has an intrinsic persuasion strength which may vary between sides (e.g. a discussion of innocent convicts may intrinsically help the anti-death-penalty side more than the pro). 
Thus we have a {\it topic strength system} $\mathbf{h_{i}=\{\mathbf{h_{i}^{p}},\mathbf{h_{i}^{c}} \}}$, where the strengths for $K$ topics are $\mathbf{h_{i}^{p}}=\{h_{i,k}^{p}\}_{k=1}^{K}$ for \textsc{pro}, and $\mathbf{h_{i}^{c}}=\{h_{i,k}^{c}\}_{k=1}^{K}$ for \textsc{con}. 
Topic strength $h_{\ast, \ast}^{\ast}$ is a categorical variable in $\{\textsc{strong}, \textsc{weak}\}$.\footnote{Binary topic strength is better-suited for our proposed discriminative learning framework. In exploratory work we found that continuous-value strength under the same framework tended to be pushed towards extreme values during learning.} 
Neither the topics nor their strength are known \textit{a priori}, and thus must be inferred.

For debate $d_{i}$, we define $\Phi (\mathbf{x_{i}^{p}}, \mathbf{h_{i}})$ and $\Phi (\mathbf{x_{i}^{c}}, \mathbf{h_{i}})$ to be feature vectors for arguments from \textsc{pro} and \textsc{con}. 
We first model and characterize features for each argument and then aggregate them by side to predict the \textit{relative success} of each side. Therefore, the feature vectors for a side can be formulated as the summation of feature vectors of its arguments, i.e. $\Phi (\mathbf{x_{i}^{p}}, \mathbf{h_{i}})=\sum_{x_{i,j} \in \mathbf{x_{i}^{p}} } \phi (x_{i, j}, \mathbf{h_{i}})$, and $\Phi (\mathbf{x_{i}^{c}}, \mathbf{h_{i}})=\sum_{x_{i,j} \in \mathbf{x_{i}^{c}} } \phi (x_{i, j}, \mathbf{h_{i}})$, where $\phi (x_{i, j}, \mathbf{h_{i}})$ is the feature vector of argument $x_{i, j}$.\footnote{This assumes the strength of arguments is additive, though it is possible that a single extremely strong or weak argument could decide a debate, or that debates are won via ``rounds'' rather than in aggregate.} 

Each argument feature in $\phi (x_{i, j}, \mathbf{h_{i}})$ combines a stylistic feature directly observed from the text with a latent strength dependent on the topic of the argument. For instance, consider an argument $x_{i, j}$ of a topic with an inferred strength of \textsc{strong} and which contains $3$ usages of the word ``you". 
Then $x_{i, j}$ has two coupled topic-aware features of the form $\phi_{M(\mathtt{feature}, \mathtt{strength})} (x_{i, j}, \mathbf{h_{i}})$: $\phi_{M(``\# you", ``strong")} (x_{i, j},  \mathbf{h_{i}})$ takes a value of 3, and $\phi_{M(``\# you", ``weak")} (x_{i, j}, \mathbf{h_{i}})$ is 0. 
$x_{i, j}$ also has a feature without strength, i.e. $\phi_{M(``\# you")} (x_{i, j},  \mathbf{h_{i}})$ = $3$. Function $M(\cdot)$ maps each feature to a unique index.

For predicting the outcome of debate $\mathbf{x_{i}}$, we compute the difference of feature vectors from \textsc{pro} and \textsc{con} in two ways: 
$\tilde{\Phi}^{p} (\mathbf{x_{i}},\mathbf{h_{i}} )= \Phi (\mathbf{x_{i}^{p}}, \mathbf{h_{i}}) - \Phi (\mathbf{x_{i}^{c}}, \mathbf{h_{i}})$ and $\tilde{\Phi}^{c} (\mathbf{x_{i}},\mathbf{h_{i}})= \Phi (\mathbf{x_{i}^{c}}, \mathbf{h_{i}}) - \Phi (\mathbf{x_{i}^{p}}, \mathbf{h_{i}})$. 
Two decision scores are computed as $f^{p}(\mathbf{x_{i}})=\max_{\mathbf{h_{i}}}[\mathbf{w} \cdot \tilde{\Phi}^{p} (\mathbf{x_{i}},\mathbf{h_{i}})]$ and $f^{c}(\mathbf{x_{i}})=\max_{\mathbf{h_{i}}}[\mathbf{w} \cdot \tilde{\Phi}^{c} (\mathbf{x_{i}},\mathbf{h_{i}})]$. The output is $1$ if $f^{p}(\mathbf{x_{i}})>f^{c}(\mathbf{x_{i}})$ (\textsc{pro} wins); otherwise, the prediction is $-1$ (\textsc{con} wins). 

Weights $\mathbf{w}$ are learned during training, while topic strengths $\mathbf{h_{i}}$ are latent variables, and we use Integer Linear Programming to search for $\mathbf{h_{i}}$. %(see \S~\ref{sec:inference}).

\subsection{Learning with Latent Variables}
\label{sec:learning}

%\vspace{-1mm}
%\paragraph{Parameter learning.} 
To learn the weight vector $\mathbf{w}$, we use the large margin training objective: 
{\small
\begin{equation}
\min_{\mathbf{w}} \frac{1}{2} \Vert \mathbf{w} \Vert ^{2} + C\cdot \sum_{i} l(-y_{i}\cdot \max_{\mathbf{h_{i}}}[\mathbf{w} \cdot \tilde{\Phi} (\mathbf{x_{i}},\mathbf{h_{i}})])
\label{eq:objective}
\end{equation}
}

\vspace{-5mm}
%The first term enforces sparsity, while the second term minimizes error.
%Intuitively, the training process can be interpreted as learning a scorer that can catch signals from the difference of two sides that leading to a winning decision. 
We consider samples based on difference feature vectors $\tilde{\Phi}^{p} (\mathbf{x_{i}},\mathbf{h_{i}} )$ during training, which is represented as $\tilde{\Phi} (\mathbf{x_{i}},\mathbf{h_{i}})$ in Eq.~\ref{eq:objective} and the rest of this section. 
%\footnote{We have a balanced dataset of positive and negative samples. For an unbalanced dataset, different strategies can be used, e.g. flipping two sides for different labels.}  
%For simplicity purpose, they are represented as $\tilde{\Phi} (\mathbf{x_{i}},\mathbf{h_{i}})$ in this section.
%
$l(\cdot)$ is squared-hinge loss function. $C$ controls the trade-off between the two items.  

This objective function is non-convex due to the maximum operation~\cite{Yu:2009:LSS:1553374.1553523}. We utilize Alg.~\ref{alg:learning}, which is an iterative optimization algorithm, to search for the solution for $\mathbf{w}$ and $\mathbf{h}$. We first initialize latent topic strength variables as $\mathbf{h_{0}}$ (see next paragraph) and learn the weight vector as $\mathbf{w^{\ast}}$.
Adopted from~\newcite{Chang:2010:DLO:1857999.1858065}, our iterative algorithm consists of two major steps. For each iteration, the algorithm first decides the latent variables for positive examples. In the second step, the solver iteratively searches for latent variable assignments for negative samples and updates the weight vector $\mathbf{w}$ with a cutting plane algorithm until convergence. Global variable $H_{i}$ is maintained for each negative sample to store all the topic strength assignments that give the highest scores during training.\footnote{A similar latent variable model is presented in~\newcite{goldwasser2014object} to predict the objection behavior in courtroom dialogues. In their work, a binary latent variable is designed to indicate the latent relevance of each utterance to an objection, and only relevant utterances contribute to the final objection decision. In our case our latent variables model argument strength, and all arguments matter for the debate outcome.} 
%
%This strategy restricts the search space to facilitate efficient training while a local optimum is guaranteed.
This strategy facilitates efficient training while a local optimum is guaranteed.

\begin{algorithm}[t]
\setstretch{0.0}
\fontsize{8.5}{9}\selectfont
\SetKwInOut{Input}{Input}
\SetKwInOut{Output}{Output}
\Input{$\{\mathbf{x_{i}}, y_{i}\}_{i}$: training samples of arguments $\mathbf{x_{i}}$ and outcome $y_{i}$, $\tilde{\Phi}(\cdot, \cdot)$: feature vectors, 
$C$: trade-off coefficient, 
$\tau$: iteration number threshold}
\Output{feature weights $\mathbf{w^{\ast}}$}
\BlankLine

%\tcp{Initialize latent topic strength}
\ForEach{$\mathbf{h_{i}}$}{
Initialize $\mathbf{h_{i}}$ as $\mathbf{h_{i}^{0}}$ (see \S~\ref{sec:learning})\;
}
%For each $i$, initialize $\mathbf{h_{i}}$ as $\mathbf{h_{i}^{0}}$ (see \S~\ref{sec:learning})\;

\;
%\tcp{Initialize weight vector}
$\mathbf{w^{\ast}} \leftarrow \arg \min_{\mathbf{w}} \frac{1}{2} \Vert \mathbf{w} \Vert ^{2} + C\cdot \sum_{i} l(-y_{i}\cdot [\mathbf{w} \cdot \tilde{\Phi} (\mathbf{x_{i}},\mathbf{h_{i}^{0}})])$\;

\;
\tcp{\fontsize{8}{9}\selectfont $H_{i}$: storing $h_{i}^{*}$ for negative samples}
\ForEach{\normalfont negative sample $\mathbf{x_{i}}$ ($y_{i}=-1$)}{
$H_{i}\leftarrow \emptyset$\;
}

\;
%\tcp{Iterative training starts}
$t \leftarrow 0$\;
\While{$w^{\ast}$ {\normalfont not converge and}  $t < \tau$}{
	
	\tcp{\fontsize{8}{9}\selectfont Assign strength for positive samples}
	\ForEach{ \normalfont positive sample $\mathbf{x_{i}}$ ($y_{i}=1$)}{
		$\mathbf{h_{i}^{\ast}} \leftarrow \arg \max_{\mathbf{h_{i}}} [\mathbf{w} \cdot \tilde{\Phi} (\mathbf{x_{i}},\mathbf{h_{i}})]$ (*)\;
	}
	
	\tcp{\fontsize{8}{9}\selectfont Iteration over negative samples}
	$t' \leftarrow 0$\;
	\While{$w^{\ast}$ {\normalfont not converge and}  $t'< \tau$}{
		\ForEach{ \normalfont negative sample $\mathbf{x_{i}}, y_{i}=-1$}{
		$\mathbf{h_{i}^{\ast}} \leftarrow \arg \max_{\mathbf{h_{i}}} [\mathbf{w^{\ast}} \cdot \tilde{\Phi} (\mathbf{x_{i}},\mathbf{h_{i}})]$ (*)\;
		
		$H_{i}\leftarrow H_{i} \cup \{\mathbf{h_{i}^{\ast}}\}$\;
		}
		
		{\fontsize{8}{9}\selectfont
		$\mathbf{w^{\ast}} \leftarrow \arg \min_{\mathbf{w}} \frac{1}{2} \Vert \mathbf{w} \Vert ^{2} 
		+ C\cdot \sum_{i,y_{i}=1} l(-y_{i}\cdot [\mathbf{w} \cdot \tilde{\Phi} (\mathbf{x_{i}},\mathbf{h_{i}^{\ast}})]) 
		+ C\cdot \sum_{i,y_{i}=-1} l(-y_{i}\cdot \max_{\mathbf{h}\in H_{i}}[\mathbf{w} \cdot \tilde{\Phi} (\mathbf{x_{i}},\mathbf{h_{i}})])$\;
		$t' \leftarrow t'+1$\;}
		
	}
	$t \leftarrow t+1$\;
	
}

\caption{\small Iterative algorithm for learning weights $\mathbf{w}$ and latent topic strength variables $\mathbf{h}$. Iteration threshold $\tau$ is set as 1000. Steps with (*) are solved as in \S~\ref{sec:inference}.}
\label{alg:learning}
\end{algorithm}

\noindent \textbf{Topic strength initialization.} 
We investigate three approaches for initializing topic strength variables. The first is based on the {\it usage frequency per topic}. If one side uses more arguments of a given topic, then its strength is likely to be strong for them and weak for their opponent. Another option is to initialize {\it all topics as strong for both sides}, then $\mathbf{w_{0}}$ learns the association between strong topics and features that lead to winning. The third option is to initialize {\it all topics as strong for winners} and weak for losers.

\subsection{Features}
\label{sec:features}

%Different categories of linguistic features are designed to characterize arguments according to their style, discourse structure, emotions, interaction with opponents, and many other aspects. 
We group our directly observed linguistic features, roughly ordered by increasing complexity, into categories that characterize various aspects of arguments. For each linguistic feature, we compute two versions: one for the full debate and one for the discussion phase.

\noindent \textbf{Basic features.} %We start with a set of basic features that describe the content, syntactic structure, and sentiment and emotional aspects of arguments. 
We consider the frequencies of words, numbers, named entities per type, and each personal pronoun. For instance, usage of personal pronouns may imply communicative goals of the speaker~\cite{brown1960pronouns,wilson1990politically}. 
We also count the frequency of each POS tag output by Stanford parser~\cite{Klein:2003:AUP:1075096.1075150}.
%\footnote{POS tagging and named entity recognition results are achieved by using Stanford parser~\cite{Klein:2003:AUP:1075096.1075150}.} 
Sentiment and emotional language usage is prevalent in discussions on controversial topics~\cite{wang-cardie:2014:P14-2}. We thus count words of positive and negative sentiment based on MPQA lexicon~\cite{Wilson:2005:RCP}, and words per emotion type according to a lexicon from \newcite{Mohammad13}. Moreover, based on the intuition that agreement carries indications on topical alignment~\cite{bender2011annotating,wang-cardie:2014:W14-26}, occurrence of agreement phrases (``I/we agree", ``you're right") is calculated. Finally, audience feedback, including applause and laughter, is also considered.

\noindent \textbf{Style features.} Existing work suggests that formality can reveal speakers' opinions or intentions~\cite{irvine1979formality}. Here we utilize a formality lexicon collected by~\newcite{Brooke:2010:AAL:1944566.1944577}, which counts the frequencies of formal or informal words in each argument. 
According to \newcite{durik2008effects}, hedges are indicators of weak arguments, so we compile a list of hedge word from~\newcite{metadiscourse10616}, and hedging of verbs and non-verbs are counted separately. 
Lastly, we measure word attributes for their concreteness (perceptible vs. conceptual), valence (or pleasantness), arousal (or intensity of emotion), and dominance (or degree of control) based on the lexicons collected by~\newcite{brysbaert2014concreteness} and~\newcite{warriner2013norms}, following~\newcite{tan+etal:16a}, who observe correlations between word attributes and their persuasive effect in online arguments. The average score for each of these features is then computed for each argument. 

\noindent \textbf{Semantic features.} We encode semantic information via semantic frames~\cite{fillmore1976frame}, which represent the context of word meanings. 
\newcite{canobasave-he:2016:N16-1} show that arguments of different types tend to employ different semantic frames, e.g., frames of ``reason" and ``evaluative comparison" are frequently used in making claims. 
We count the frequency of each frame, as labeled by SEMAFOR~\cite{das2014frame}.

\noindent \textbf{Discourse features.} The usage of discourse connectors has been shown to be effective for detecting argumentative structure in essays~\cite{stab2014identifying}. We collect discourse connectives from the Penn Discourse Treebank~\cite{prasad2007penn}, and count the frequency of phrases for each discourse class. Four classes on level one (temporal, comparison, contingency, and expansion) and sixteen classes on level two are considered. 
%\footnote{We consider four discourse classes on level one (temporal, comparison, contingency, and expansion), and sixteen classes on level two.}
% (Asynhronous, Synhronous, Contrast, Pragmati Contrast, Conession, Pragmati Conession, Cause, Pragmati Cause, Condition, Pragmati Condition, Conjuntion, Instantiation, Restatement, Alternative, Exeption, and List).}
Finally, pleading behavior is encoded as counting phrases of ``urge", ``please", ``ask you", and ``encourage you", which may be used by debaters to appeal to the audience.

\noindent \textbf{Sentence-level features.} 
We first consider the frequency of questioning since rhetorical questions are commonly used for debates and argumentation. 
To model the sentiment distribution of arguments, sentence-level sentiment is labeled by the Stanford sentiment classifier~\cite{socher-EtAl:2013:EMNLP} as positive (sentiment score of 4 or 5), negative (score of 1 or 2), and neutral (score of 3). We then count single sentence sentiment as well as transitions between adjacent sentences (e.g. positive $\rightarrow$ negative) for each type. 
Since readability level may affect how the audience perceives arguments, we compute readability levels based on Flesch reading ease scores, Flesch-Kincaid grade levels, and the Coleman-Liau index for each sentence. We use the maximum, minimum, and average of scores as the final features. 
The raw number of sentences is also calculated.

\noindent \textbf{Argument-level features.} 
Speakers generally do not just repeat their best argument ad infinitum, which suggests that arguments may lose power with repetition. For each argument, we add an indicator feature (i.e. each argument takes value of $1$) and an additional version with a decay factor of $\exp (-\alpha \cdot t_{k})$, where $t_{k}$ is the number of preceding arguments by a given side which used topic $k$, and $\alpha$ is fixed at $1.0$. %\footnote{For future work, we aim to automatically learn $\alpha$.}
Interruption is also measured, when the last argument (of more than 50 words) in a turn is cut off by at most two sentences from opponent or moderator. 
Word repetition is often used for emphasis in arguments~\cite{canobasave-he:2016:N16-1}, so we measure bigram repetition more than twice in sequential clauses or sentences. 
%We then consider alliteration by detecting bigram repetition more than twice in sequential clauses or sentences.

\noindent \textbf{Interaction features.} 
In addition to independent language usage, debate strategies are also shaped by interactions with other debaters. For instance, previous work~\cite{zhangetal:NAACL2016} finds that debate winners frequently pursue talking points brought up by their opponents'. 
Here we construct different types of features to measure how debaters address opponents' arguments and shift to their favorable subjects. 
First, for a given argument, we detect if there is an argument of the same topic from the previous turn by the opponent. If yes, we further measure the number of words of the current argument, the number of common words between the two arguments (after lemmatization is applied), the concatenation of the sentiment labels, and the concatenation of the emotion labels of the two arguments as features; these interactions thus capture interactive strategies regarding quantity speech and sentiment.
We also consider if the current argument is of a different topic from the previous argument in the same turn to encode topic shifting behavior. 

Feature functions $\phi_{M(\mathtt{feature}, \mathtt{strength})} (x_{i, j}, \mathbf{h_{i}})$ in \S~\ref{sec:problem} only consider the strengths of single arguments. To capture interactions between sides that relate to their relative argument strengths, we add features $\phi_{M(\mathtt{feature}, \mathtt{strength^{self},strength^{oppo}})} (x_{i, j}, \mathbf{h_{i}})$, so that strengths of pairwise arguments on the same topic from both sides are included. 
For instance, for topic ``execution of innocents", side \textsc{pro} with \textsc{strong} strength uses an argument of 100 words to address the challenge raised by \textsc{con} with \textsc{weak} strength. We add four grouped features associated with the number of words addressing an opponent: $\phi_{M(``\# words~to~oppo",``strong,weak")} (x_{i, j}, \mathbf{h_{i}})$ is 100, while $\phi_{M(``\# words~to~oppo", ``weak,weak")} (x_{i, j}, \mathbf{h_{i}})$, $\phi_{M(``\# words~to~oppo", ``strong,strong")} (x_{i, j}, \mathbf{h_{i}})$, and $\phi_{M(``\# words~to~oppo",``weak,strong")} (x_{i, j}, \mathbf{h_{i}})$ are all 0. 

%\subsubsection{Adding Topic Strength to Features}
%As presented in \S~\ref{sec:problem}, our features $\phi (x_{i, j}, h_{z_{i, j}})$ model the interaction between language usage and the intrinsic topic strength for each argument. 
%Given an argument $x$, let $z_{x}$ be the topic of $x$. We consider two types of feature functions: \textbf{local features} $\phi (x)$ that only consider information from a single argument, and \textbf{interaction features} $\phi (x,x^{\prime})$ that characterize interplay between arguments from different sides. 

%In order to enable our model to learn the interaction between language usage and argument's topic strength, we also create \textbf{topic strength-specific} versions as {\small $\phi (x, h_{z_{x}})$} and {\small $\phi (x, x^{\prime}, h_{z_{x}})$} for local features, and {\small $\phi (x,x^{\prime},h_{z_{x}},h_{z_{x^{\prime}}})$} for interaction features.
%For each feature, we compute two versions: one for the full debate, the other for the discussion phase.

\subsection{Topic Strength Inference}
\label{sec:inference}
Topic strength inference is used both for training (Alg.~\ref{alg:learning}) and for prediction. 
Our goal is to find an assignment $\mathbf{h_{i}^{*}}$ that maximizes the scorer $\mathbf{w^{*}} \cdot \tilde{\Phi} (\mathbf{x_{i}},\mathbf{h_{i}})$ for a given $\mathbf{w^{*}}$. 
We formulate this problem as an Integer Linear Programming (ILP) instance.\footnote{We use LP Solve: \url{http://lpsolve.sourceforge.net/5.5/}.} Since topic strength assignment only affects feature functions that consider strengths, here we discuss how to transform those functions into the ILP formulation.

%Consider feature functions $\phi_{M(\mathtt{feature}, \mathtt{strength})} (x^{p}, h_{z^{p}})$ and $\phi_{M(\mathtt{feature}, \mathtt{strength})} (x^{c}, h_{z^{c}})$, where $x^{p}$ and $x^{c}$ are arguments from pro and con ($z^{p}$ and $z^{c}$ are corresponding topic indice).

For each topic $k$ of a debate $d_{i}$, we create binary variables $r_{k, strong}^{p}$ and $r_{k, weak}^{p}$ for pro, where $r_{k, strong}^{p} = 1$ indicates the topic is \textsc{strong} for pro and $r_{k, weak}^{p}=1$  denotes the topic is \textsc{weak}. Similarly, $r_{k, strong}^{c}$ and $r_{k, weak}^{c}$ are created for con. 

%For any feature, remember that we have learned weights $w_{M(\mathtt{feature}, strong)}$ and $w_{M(\mathtt{feature}, weak)}$ associated with different topic strengths. 
Given weights associated with different strengths $w_{M(\mathtt{feature}, strong)}$ and $w_{M(\mathtt{feature}, weak)}$, the contribution of any feature to the objective (i.e. scoring difference between pro and con) transforms from 

%\vspace{-4mm}
%{\fontsize{9}{9}\selectfont
%\begin{equation*}
%\begin{split}
%&w_{M(\mathtt{feature}, strong)}\cdot [\sum_{x_{i,j} \in \mathbf{x_{i}^{p}}} \phi_{M(\mathtt{feature}, strong)} (x_{i,j}, \mathbf{h_{i}}) \\
%& - \sum_{x_{i,j} \in \mathbf{x_{i}^{c}}} \phi_{M(\mathtt{feature}, strong)} (x_{i,j}, \mathbf{h_{i}})] \\
%&+w_{M(\mathtt{feature}, weak)}\cdot [\sum_{x_{i,j} \in \mathbf{x_{i}^{p}}} \phi_{M(\mathtt{feature}, weak)} (x_{i,j}, \mathbf{h_{i}}) \\
%& - \sum_{x_{i,j} \in \mathbf{x_{i}^{c}}} \phi_{M(\mathtt{feature}, weak)} (x_{i,j}, \mathbf{h_{i}})] \\
%\end{split}
%\end{equation*}
%}

\vspace{-1mm}
\noindent
{\fontsize{8.5}{8}\selectfont
$w_{M(\mathtt{feature}, strong)}\cdot [\sum_{x_{i,j} \in \mathbf{x_{i}^{p}}} \phi_{M(\mathtt{feature}, strong)} (x_{i,j}, \mathbf{h_{i}}) \\
 - \sum_{x_{i,j} \in \mathbf{x_{i}^{c}}} \phi_{M(\mathtt{feature}, strong)} (x_{i,j}, \mathbf{h_{i}})] \\
+w_{M(\mathtt{feature}, weak)}\cdot [\sum_{x_{i,j} \in \mathbf{x_{i}^{p}}} \phi_{M(\mathtt{feature}, weak)} (x_{i,j}, \mathbf{h_{i}}) \\
 - \sum_{x_{i,j} \in \mathbf{x_{i}^{c}}} \phi_{M(\mathtt{feature}, weak)} (x_{i,j}, \mathbf{h_{i}})] \\
$
}

\vspace{-7mm}
to the following form:

\vspace{-1mm}
\noindent
{\fontsize{8.5}{8}\selectfont
$w_{M(\mathtt{feature}, strong)}\cdot [\sum_{x_{i,j} \in \mathbf{x_{i}^{p}}} \phi_{M(\mathtt{feature})} (x_{i,j}, \mathbf{h_{i}})\cdot r_{k,strong}^{p} \\
 - \sum_{x_{i,j} \in \mathbf{x_{i}^{c}}} \phi_{M(\mathtt{feature})} (x_{i,j}, \mathbf{h_{i}})\cdot r_{k,strong}^{c} ] \\
+w_{M(\mathtt{feature}, weak)}\cdot [\sum_{x_{i,j} \in \mathbf{x_{i}^{p}}} \phi_{M(\mathtt{feature})} (x_{i,j}, \mathbf{h_{i}})\cdot r_{k,weak}^{p} \\
 - \sum_{x_{i,j} \in \mathbf{x_{i}^{c}}} \phi_{M(\mathtt{feature})} (x_{i,j}, \mathbf{h_{i}})\cdot r_{k,weak}^{c} ] \\
$
}

%\vspace{-3mm}
%{\fontsize{9}{9}\selectfont
%\begin{equation*}
%\begin{split}
%&w_{M(\mathtt{feature}, strong)}\cdot [\sum_{x_{i,j} \in \mathbf{x_{i}^{p}}} \phi_{M(\mathtt{feature})} (x_{i,j}, \mathbf{h_{i}})\cdot r_{k,strong}^{p} \\
%& - \sum_{x_{i,j} \in \mathbf{x_{i}^{c}}} \phi_{M(\mathtt{feature})} (x_{i,j}, \mathbf{h_{i}})\cdot r_{k,strong}^{c} ] \\
%&+w_{M(\mathtt{feature}, weak)}\cdot [\sum_{x_{i,j} \in \mathbf{x_{i}^{p}}} \phi_{M(\mathtt{feature})} (x_{i,j}, \mathbf{h_{i}})\cdot r_{k,weak}^{p} \\
%& - \sum_{x_{i,j} \in \mathbf{x_{i}^{c}}} \phi_{M(\mathtt{feature})} (x_{i,j}, \mathbf{h_{i}})\cdot r_{k,weak}^{c} ] \\
%\end{split}
%\label{eq:ilp}
%\end{equation*}
%}

\vspace{-5mm}
The above equation can be reorganized into a linear combination of variables $r_{*, *}^{*}$. We further include constraints as discussed below, and solve the maximization problem as an ILP instance.  %Remember that weights $w_{M(\mathtt{feature}, strong)}$ and $w_{M(\mathtt{feature}, weak)}$ associated with different topic strengths are given. %If $r_{strong}^{p}$ is $1$, then the topic strength is strong for pro; if  $r_{weak}^{p}$ is $1$, then it is weak. Similar results apply for con.

For features that consider strength for pairwise arguments, i.e. $\phi_{M(\mathtt{feature}, \mathtt{strength^{self},strength^{oppo}})}$, we have binary variables $r_{k, strong, strong}^{p, c}$ (strength is strong for both sides), $r_{k, strong, weak}^{p, c}$ (strong for pro, weak for con), $r_{k, weak, strong}^{p, c}$ (weak for pro, strong for con), and $r_{k, weak, weak}^{p, c}$ (weak for both).

%For each topic $k$, we create binary variables {\small $h_{k}^{p,s}$} and {\small $h_{k}^{p,w}$} for pro, where {\small $h_{k}^{p,s}$} indicates whether topic $k$ is \textsc{strong} for pro, and {\small $h_{k}^{p,w}$} denotes whether it is \textsc{weak} for pro; similarly, we have {\small $h_{k}^{c,s}$} and {\small $h_{k}^{c,w}$} for con. With this notation, feature {\small $\phi (x^{p}, h_{k})$} for pro, where $x^{p}$ is of topic $k$, can be expressed as {\small $\phi_{s} (x^{p}) \cdot h_{k}^{p,s}$} and {\small $\phi_{w} (x^{p}) \cdot h_{k}^{p,w}$}, where $\phi_{s} (\cdot)$ and $\phi_{w} (\cdot)$ are features associated with topic of strong or weak. 

%For interaction features {\small $\phi (x^{p},x^{c},h_{k},h_{m})$}, in addition to {\small $h_{k}^{p,s}$, $h_{k}^{p,w}$, $h_{m}^{c,s}$, $h_{m}^{c,w}$}, we create interaction binary variables {\small $h_{k,m}^{p,c,s,s}$, $h_{k,m}^{p,c,s,w}$, $h_{k,m}^{p,c,w,s}$}, and $h_{k,m}^{p,c,w,w}$ to encode the topic strength for both arguments. 

\noindent \textbf{Constraints.} We consider three types of topic strength constraints for our ILP formulation:

\noindent $\bullet$ C1 -- \textit{Single Topic Consistency}: each topic can either be strong or weak for a given side, but not both.\\
pro: {\fontsize{9}{9}\selectfont $r_{k, strong}^{p}+r_{k, weak}^{p}=1$}; 
con: {\fontsize{9}{9}\selectfont $r_{k, strong}^{c}+r_{k, weak}^{c}=1$}

\noindent $\bullet$ C2 -- \textit{Pairwise Topic Consistency}: for pairwise arguments from pro and con on the same topic, their joint assignment is true only when each of the individual assignments is true. C2 applies only for features of pairwise arguments.\\
{\small $r_{k, strong, strong}^{p, c}= r_{k, strong}^{p} \wedge r_{k, strong}^{c} $};\\
{\small $r_{k, strong, weak}^{p, c}= r_{k, strong}^{p} \wedge r_{k, weak}^{c} $};\\
{\small $r_{k, weak, strong}^{p, c}= r_{k, weak}^{p} \wedge r_{k, strong}^{c} $};\\
{\small $r_{k, weak, weak}^{p, c}= r_{k, weak}^{p} \wedge r_{k, weak}^{c} $}

\noindent $\bullet$ C3 -- \textit{Exclusive Strength}: a topic cannot be strong for both sides. This constraint is optional and will be tested in experiments. {\small $r_{k, strong}^{p}+r_{k, strong}^{c}\leq 1$}

\subsection{Argument Identification}
\label{sec:argextraction}
In order to identify the topics associated with a debate and the contiguous chunks of same-topic text that we take to be arguments, for each separate debate we utilize a Hidden Topic Markov Model (HTMM)~\cite{gruber2007hidden} which jointly models the topics and topic transitions between sentences. For details on HTMM, we refer the readers to~\newcite{gruber2007hidden}.

The HTMM assigns topics on the sentence level, assuming each sentence is generated by a topic draw followed by word draws from that topic, with a transition probability determining whether each subsequent sentence has the same topic as the preceding, or is a fresh draw from the topic distribution. Unlike the standard HTMM process, however, we presume that while both sides of a debate share the same topics, they may have different topic distributions reflecting the different strengths of those topics for either side. We thus extend the HTMM by allowing different topics distributions for the pro and con speech transcripts, but enforce shared word distributions for those topics. To implement this, we first run HTMM on the entire debate, and then rerun it on the pro and con sides while fixing the topic-word distributions. Consecutive sentences by the same side with the same topic are treated as a single argument. %In future work, we will enhance our argument extraction component by including interaction information. 
%\footnote{In future work, we will enhance our argument extraction component by including interaction information.}

%% file: prediction.tex
\subsection{Experimental Setup}
We test via leave-one-out for all experiments. For logistic regression classifiers, L2 regularization with a trade-off parameter of 1.0 is used. For Support Vector Machines (SVM) classifiers and our models, we fix the trade-off parameter between training error and margin as 0.01. Real-valued features are normalized to $[0, 1]$ via linear transformation.

Our modified HTMM is run on each debate for between 10 and 20 topics. Topic coherence, measured via~\newcite{Roder:2015:EST:2684822.2685324}, is used to select the topic number that yields highest score. On average, there are 13.7 unique topics and about 322.0 arguments per debate.

\subsection{Baselines and Comparisons}
We consider two baselines trained with logistic regression and SVMs classifiers: (1) \textsc{ngrams}, including unigrams and bigrams, are used as features, and (2) \textsc{audience feedback} (applause and laughter) are used as features, following \newcite{zhangetal:NAACL2016}. 
We also experiment with SVMs trained with different sets of features, presented in \S~\ref{sec:features}. 
%
%\newcite{zhangetal:NAACL2016} studies the usage of talking points (i.e. sets of key words) by opposing sides, and tests if the difference between debaters carries predictive power. We also compare with this approach. 

\subsection{Results}

The debate outcome prediction results are shown in Table~\ref{tab:prediction}. For our models, we only display results with latent strength initialization based on frequency per topic, which achieves the best performance. Results for different initialization methods are exhibited and discussed later in this section. 
As can be seen, our model that leverages learned latent topic strengths and their interactions with linguistic features significantly outperform the non-trivial baselines\footnote{For baselines with logistic regression classifiers, the accuracy is 63.6 with ngram features, and 58.5 with audience feedback features.} 
(bootstrap resampling test, $p<0.05$). 
Our latent variable models also obtain better accuracies than SVMs trained on the same linguistic feature sets. Without the audience feedback features, our model yields an accuracy of 72.0\%, while SVM produces 65.3\%.
%\footnote{Without the audience feedback features, our model yields an accuracy of 72.0\%, while SVM classifier obtains 65.3\%. }  
This is because our model can predict topic strength out of sample by learning the interaction between observed linguistic features and unobserved latent strengths. 
During test time, it infers the latent strengths of entirely new topics based on observable linguistic features, and thereby predict debate outcomes more accurately than using the directly observable features alone. 
Using the data in \newcite{zhangetal:NAACL2016} (a subset of ours),
our best model obtains an accuracy of 73\% compared to 65\% based on leave-one-out setup.
%Our model outperforms their approach based on distinguishing talking points among debaters, which achieved 65\% for accuracy.

\begin{table}[t]
    {\fontsize{9}{10}\selectfont

    \setlength{\tabcolsep}{1.5mm}
	\vspace{2mm}
	\begin{tabular}{l c c}
	\hline
	& {\bf SVMs} & {\bf Our Model}\\ 
	&  & ({\it w Latent}\\ 
		&  & {\it Strength})\\ 
	\underline{\bf Baselines}	 & & \\ 
    \textsc{Ngrams} & 61.0 & --\\
    \textsc{Audience Feedback} & 56.8 & --\\	

	\\
	\underline{\bf Features} (as in \S~\ref{sec:features}) & & \\
	\textsc{Basic} 							&	57.6		&	59.3		\\
    + \textsc{Style, Semantics, Discourse} 	&	59.3		&	65.3		\\
	+ \textsc{Sentence, Argument} 			&	62.7		&	69.5		\\
	+ \textsc{Interaction} 	(all features)				&	66.1		&	\textbf{73.7}	\\
	
    \hline
	\end{tabular}
	}
	\vspace{-3mm}
    \caption{\small Debate outcome prediction results for baseline models and SVMs using the various linguistic feature categories, compared to our model that includes latent argument strengths in addition to the linguistic features. The best performing system (in \textbf{bold}) is achieved by our system with topic strength as latent variables when all features are used, which significantly outperforms the baselines via bootstrap resampling test ($p<0.05$). For the lower section, each row shows features included in addition to those in the rows above.
    }
    
	\label{tab:prediction}
\end{table}

As mentioned above, we experimented with a variety of latent topic strength initializations: argument frequency per topic (\textit{Freq}); all topics strong for both sides (\textit{AllStrong}); strong just for winners (\textit{AllStrong$_{win}$}); and \textit{Random} initialization. 
%Results on using different initialization methods and different sets of topic strength constraints are displayed in Table~\ref{tab:prediction_init}. 
From Table~\ref{tab:prediction_init}, we can see that there is no significant difference among different initialization methods. 
Furthermore, the strength constraints make little difference, though their effects slightly vary with different initializations. Most importantly, C3 (the constraint that topics cannot be strong for both sides) does not systematically help, suggesting that in many cases topics may indeed be strong for both sides, as discussed below.

\vspace{-3mm}
\begin{table}[ht]
\centering
    {\fontsize{10}{10}\selectfont
    \setlength{\tabcolsep}{1.0mm}

    \begin{tabular}{l c c c c}
    \hline
									& \multicolumn{4}{c}{\bf Initialization}\\
	\textit{Constraints}				& \textit{Freq} & \textit{AllStrong} & \textit{AllStrong$_{win}$} & \textit{Random}\\
	\textsc{C1, C2 } 				& 73.7	& 71.2	& 70.3	& 67.8 	\\
	\textsc{C1, C2, C3} 			& 72.9	& 73.7	& 69.5	& 68.6	\\
    
    \hline
	\end{tabular}
	}
	\vspace{-3mm}
    \caption{\small Prediction results (in accuracy) with different initialization and topic strength constraints. \textit{C3} denotes a constraint that a topic cannot be strong for both sides.}
	\label{tab:prediction_init}
\end{table}

%% file: discussion.tex
In this section, we first analyze argument strengths for winning and losing sides, followed by a comparison of these results with human evaluations (\S~\ref{sec:humaneval}). 
We then examine the interactive topic shifting behavior of debaters (\S~\ref{sec:shifting}) and analyze the linguistic features predictive of debate outcome, particularly the ones that interact with topic strength (\S~\ref{sec:feature_analysis}). 
The results are reported by training our model on the full dataset. Initialization of topic strength is based on usage frequency unless otherwise specified.

\begin{figure}[t]
\centering
\includegraphics[width=65mm,height=45mm]{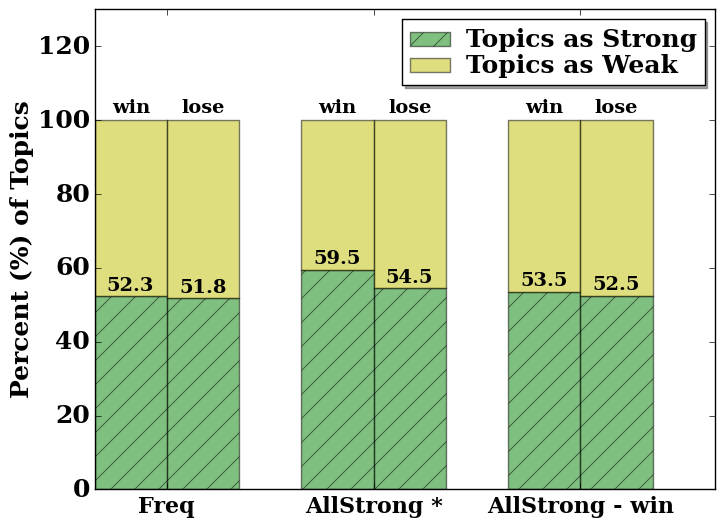}
\includegraphics[width=65mm,height=45mm]{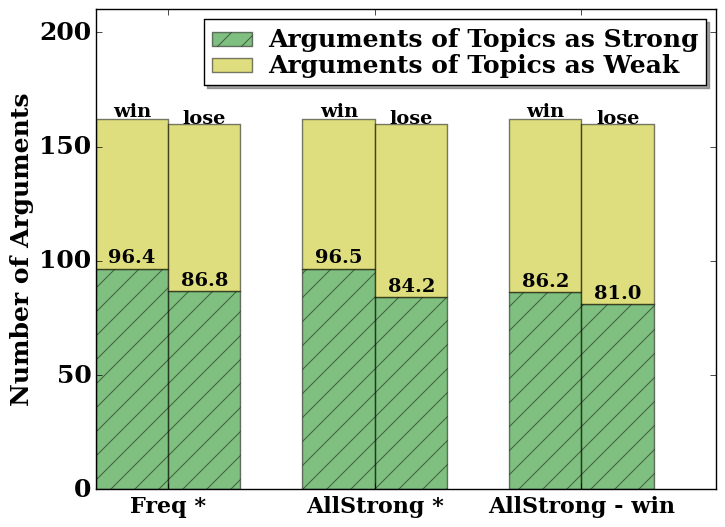}
\vspace{-3mm}
\caption{\small [Upper] Average percentage of topics inferred as \textsc{Strong} and \textsc{Weak} for winning (``win") and losing sides (``lose"). [Lower] Raw number of arguments of \textsc{Strong} and \textsc{Weak} topics. Numbers are computed for three types of topic strength initialization: initialized by frequency (Freq), all topics are strong for both sides (AllStrong), and all topics are strong for winners (AllStrong - win). 
Two-sided Mann-Whitney rank test is conducted on values of \textsc{Strong} topics ($\ast$: $p<0.05$).}
\label{fig:topicpercentage}
\end{figure}

\subsection{Topic and Argument Usage Analysis}
\label{sec:topic_analysis}

We start with a basic question: \textit{do winning sides more frequently use strong arguments?} For each side, we calculate the proportion of strong and weak topics as well as the total number of strong and weak arguments on each side. Figure~\ref{fig:topicpercentage} shows that under all three topic strength initializations, our model infers a greater number of strong topics for winners than for losers. 
%\footnote{Note that this is the same framework we use for out-of-sample prediction; by contrast, assuming that all topics used more by the winning team are strong, for instance, would be a useless model for out-of-sample prediction.}
This result is echoed by human judgment of topic strength, as described in \S~\ref{sec:humaneval}. Similarly, winners also use significantly more individually strong arguments.

\begin{figure}[ht]
	{\fontsize{9}{10}\selectfont
    \setlength{\tabcolsep}{0.6mm}
    \hspace{-2mm}
    \begin{tabular}{|p{82mm}|}
    \hline
    
    %\underline{\textbf{Motion}}: {\it Abolish The Minimum Wage} \\
	%\textbf{Topic}: Minimum wage and employment\\
	%$\bullet$ Pro (\textsc{Strong}): 
	%If the wage rate dropped for some workers, the lowest of the lower skilled workers, yeah, their wage rate may go down in the short run, but they'll have a job. 
	%And their wage rate's not going to stay at a low level forever if they work hard. That's the point. You're pricing these -- the politicians are pricing these people out of the market in the name of doing good. It's a feel-good policy, and the empirical evidence -- 
	%...the preponderance of empirical evidence supports the idea that raising the wage rate arbitrarily will reduce jobs more so in the long run. \\
	%$\bullet$ Con (\textsc{Weak}):
	%Studies suggest that other factors, such as the overall state of the economy, how local industries are doing, matter a lot more for employment than the level of the minimum wage does. \\
	
	%\hline
	%\hline
	
    \underline{\textbf{Motion}}: {\it The U.S. Should Let In 100,000 Syrian Refugees} \\
	\textbf{Topic}: Refugee resettlement\\
	$\bullet$ Pro (\textsc{Strong}): 
	415 Syrians resettled by the IRC. Our services show that last year, 8 out of 10 Syrians who we resettled were in work within six months of getting to the United States. And there's one other unique resource of this country: Syrian-American communities across the country who are successful. ... %The business people in San Diego, the doctors in Cleveland, the shopkeepers in New York -- there are Syrian-American communities ready to welcome and integrate more Syrians who come here. 
	\\
	$\bullet$ Con (\textsc{Strong}):
	It costs about 13 times as much to resettle one refugee family in the United States as it does to resettle them closer to home. ...
	They're asking you to look only at the 400 -- the examples of the 415 Syrians that David Miliband's group has so well resettled, and to ignore what is likely to happen as the population grows bigger. 
	\\	
    
    %\hline
	%\hline
	
    %\underline{\textbf{Motion}}: {\it Income %Inequality Impairs The American Dream of Upward Mobility} \\
	%\textbf{Topic}: Hard-working and tax\\
	%$\bullet$ Pro (\textsc{Weak}): 
	%Do you think you work harder than your parents because your tax rates are lower than theirs? I don't think so. So hard work is -- is an extremely important thing, but it is completely uncorrelated to the tax rates we pay.  \\
	%$\bullet$ Con (\textsc{Weak}):
	%I think it's a little bit disingenuous to take a poll and say that tax rates don't affect work. 
	%I mean, the argument that smart people on the left that I know make about why inequality at the top has risen is because of the Reagan tax cuts, right? The Reagan tax cuts -- this is the argument from the left -- increased the rewards to bargaining. 
	%...because tax cuts went down, it became more advantageous for people at the top to bargain, to take risk, to work more, and that's why income concentration has risen.
%\\	
        
    \hline
    \end{tabular}
    }
   	\vspace{-4mm}
	\caption{\small A sample exchange where the argument topic is strong for both sides.}
\label{fig:example_arguments}
\end{figure}

%\vspace{-3mm}
\begin{figure}[ht]
\centering
%\vspace{-8mm}
    \includegraphics[width=58mm,height=42mm]{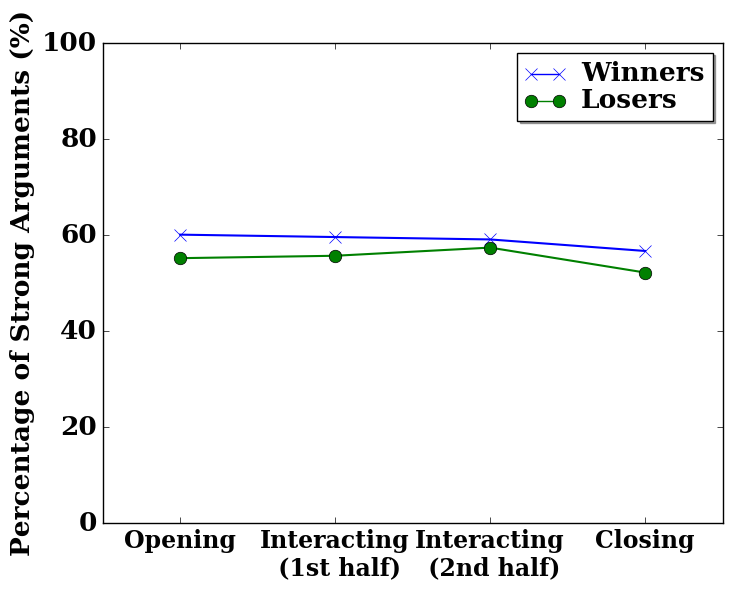}
\vspace{-4mm}
\caption{\small Usage of arguments with strong topics at different stages for winners and losers. Similar results are achieved by counting the number of words in arguments.}
\label{fig:argumentshift}
\end{figure}

%\begin{figure}[ht]
%\centering
%\vspace{-8mm}
%\subfloat
%{	
%	\hspace{-4mm}
%    \includegraphics[width=40mm,height=35mm]{figs/argumentshift_win.png}
%}
%\subfloat
%{	
%	\hspace{-3mm}
%
%    \includegraphics[width=40mm,height=35mm]{figs/argumentshift_lose.png}
%}
%%\vspace{-4mm}
%\caption{\small Usage of arguments of strong topics at different stages for winners and losers. Similar results are achieved by counting number of words in arguments.}
%\label{fig:argumentshift}
%\end{figure}

%\begin{figure}[ht]
%\centering
%%\vspace{-8mm}
%    \includegraphics[width=62mm,height=45mm]{figs/argumentshift_win.png}
%    \includegraphics[width=62mm,height=45mm]{figs/argumentshift_lose.png}
%%\vspace{-4mm}
%\caption{\small Usage of arguments of strong topics at different stages for winners and losers. Similar results are achieved by counting number of words in arguments.}
%\label{fig:argumentshift}
%\end{figure}

As can be seen in Table~\ref{tab:prediction_init}, the C3 constraint, that a topic be strong for at most one side, only increased accuracy for one initialization case. This indicates that, in general, the model was improved by allowing some topics to be strong for both sides. Interestingly, while the majority (53\%) of topics are  \textsc{strong} for one side and \textsc{weak} for the other, about a third (31\%) of topics are inferred as \textsc{strong} for both sides. While it is clear what it means for a topic to be strong for one side and not the other (as in our death penalty example), or weak for both sides (as in a digression off of the general debate topic), the importance of both-strong for prediction is a somewhat surprising result. Figure~\ref{fig:example_arguments} illustrates an example as judged by our model. What this shows is that even on a given topic within a debate (Syrian refugees: resettlement), there are different subtopics that may be selectively deployed (resettlement success; resettlement cost) that make the general topic strong for both sides in different ways. For subsequent work, a hierarchical model with nested strength relationships~\cite{mccombs2005look,nguyen-EtAl:2015:ACL-IJCNLP2} can be designed to better characterize the topics. 
%For subsequent work, this points to a hierarchical model with nested strength relationships.

Lastly, we display the usage of strong arguments during the course of debates in Figure~\ref{fig:argumentshift}. Each debate is divided into opening statements, two interacting phases (equal number of turns), and closing statements. Similar usage of strong arguments are observed as debates progress, though a slight, statistically non-significant drop is noted in the closing statement. One possible interpretation is that debaters have fully delivered their strong arguments during opening and interactions, while only weaker arguments remain when closing the debates.

%\begin{table}[th]
%\centering
%    {\fontsize{9}{10}\selectfont
%    \setlength{\baselineskip}{0pt}
%
%    \begin{tabular}{| l  l |}
%    \hline
%    \textbf{Topic Strength} & \textbf{Percent ($\pm$ STD)}\\ 
%	Strong \& Strong & 30.8\% ($\pm 0.15$) \\
%	Strong \& Weak & 52.7\% ($\pm 0.15$) \\
%	Weak \& Weak & 16.5\% ($\pm 0.11$)\\
%	\hline
%	\end{tabular}
%	}
%	\vspace{-2mm}
%    \caption{\small Proportions of topics that are assigned as Strong \& Strong, Strong \& Weak, and Weak \& Weak for different sides. }
%	\label{tab:topic_assignment}
%\end{table}

\subsection{Human Validation of Topic Strength}
\label{sec:humaneval}
Here we evaluate \textit{whether our inferred topic strength matches human judgment.} We randomly selected 20 debates with a total of 268 topics. For each debate, we first displayed its motion and a brief description constructed by IQ2. Then for each topic, the top 30 topic words from the HTMM model were listed, followed by arguments from \textsc{pro} and \textsc{con}. Note that debate results were not shown to the annotators.

We hired three human annotators who are native speakers of English. Each of them was asked to first evaluate topic coherence by reading the word list and rate it on a 1-3 scale (1 as incoherent, 3 as very coherent). If the judgment was coherent (i.e. a 2 or 3), they then read the arguments and judged whether (a) both sides are strong on the topic, (b) both sides are weak, (c) pro is strong and con is weak, or (d) con is strong and pro is weak.
%\footnote{Though annotators aimed to be non-biased in this assessment, in many cases it often came down to subjective interpretation.} 
%
%Each annotator spent about 8 hours finishing all the annotations.

54.9\% of the topics were labeled as coherent by at least two annotators, although since topics are estimated separately for each debate, even the less coherent topics generally had readily interpretable meanings in the context of a given debate. Among coherent topics, inter-annotator agreement for topic strength annotation had a Krippendorff's $\alpha$ of 0.32. Judging argument strength is clearly a more difficult and subjective task.

%There are about 54.9\% of the topics are annotated as coherent (2 or 3) by at least two annotators, and 13.1\% are annotated as coherent by all three annotators. Inter-annotator agreement based on Krippendorff's $\alpha$ is calculated as 0.32 and 0.24 for the two scenarios.

Nevertheless, without knowing debate outcomes, all three human judges identified more strong topics for winning sides than losing sides. Specially, among the coherent topics, a (macro-)average of 44.4\% of topics were labeled as strong for winners, compared to 30.1\% for losers. This echoes the results from our models as illustrated in Figure~\ref{fig:topicpercentage}.

\begin{figure}[t]
\centering
    \includegraphics[width=72mm,height=48mm]{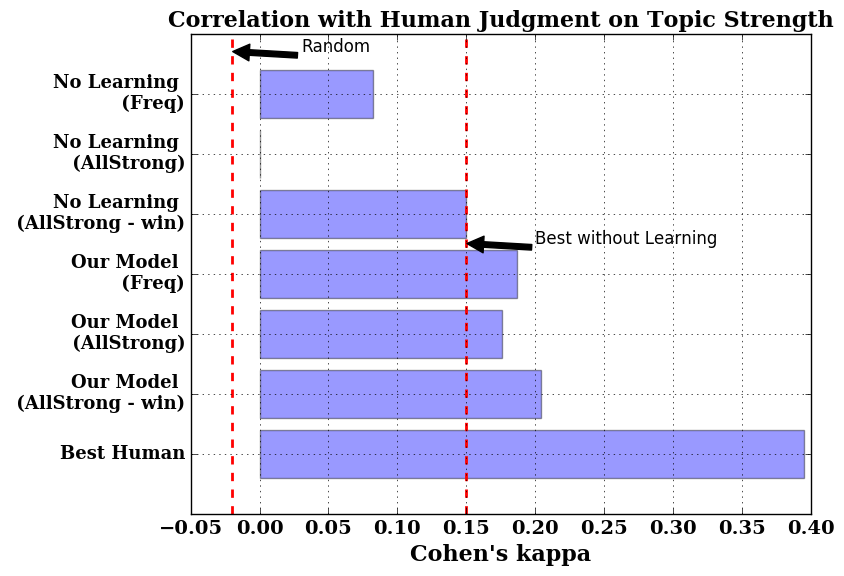}
\vspace{-4mm}
\caption{\small Topic strength correlation with human judgment using Cohen's $\kappa$. The left red dotted line indicates the best correlation between a random assignment and human, and the right red dotted line shows the best correlation without learning. %Topic strength assignments output by our models better correlate with human judgment than assignments without learning.
}
\label{fig:strengthcorrelation}
\end{figure}

Furthermore, we calculate the correlation between topic strength inferred by our models and the ones labeled by each human judge using Cohen's $\kappa$. The results are illustrated in Figure~\ref{fig:strengthcorrelation}, which shows our three different initializations, with and without learning. The highest human $\kappa$ is also displayed. 
%
%The highest human $\kappa$ is displayed for each assignment along with the best pairwise human annotation agreement.  
Our trained models clearly match human judgments better than untrained ones.

\subsection{Topic Shifting Behavior Analysis}
\label{sec:shifting}
% add more setup description
Within competitive debates, strategy can be quite interactive: one often seeks not just to make the best arguments, but to better the opponent from round to round.  Agenda setting, or shifting the topic of debate, is thus a crucial strategy.  An important question is therefore: \textit{do debaters strategically change topics to ones that benefit themselves and weaken their opponent?} 
According to the HTMM results, debaters make 1.5 topical shifts per turn on average.  %Table~\ref{tab:topic_shift} shows that 
Both winning and losing teams are more likely to change subjects to their strong topics: winners in particular are much more likely to change the topic to something strong for them (61.4\% of shifts), although debate losers also attempt this strategy (53.6\% of shifts).

% \vspace{-3mm}
% \begin{table}[th]
%     \centering

%     {\fontsize{9}{10}\selectfont
%     \setlength{\baselineskip}{0pt}

%     \begin{tabular}{ c | c c |  c c }
%     \hline
%     		& \multicolumn{2}{c|}{\textbf{Winners}} & \multicolumn{2}{c}{\textbf{Losers}}\\ 
% 	\textbf{Shift-to}   & \textsc{Strong} & \textsc{Weak} & \textsc{Strong} & \textsc{Weak}\\
% 							&	61.4\%&	38.6\%&	53.6\%&	46.4\%\\	

% %	\textsc{Strong} & 61.4\% & 53.6\% \\
% %	\textsc{Weak} & 38.6\% & 46.4\% \\
% 	\hline
% 	\end{tabular}
% 	}
% %	\vspace{-3mm}
%     \caption{\small Topic shifting behavior. Both winners and losers tend to change the subject to topics that are \textsc{strong} for their side.}
% 	\label{tab:topic_shift}
% \end{table}

\begin{table}[t]
    {\fontsize{8}{9}\selectfont
    \setlength{\baselineskip}{0pt}
    \setlength{\tabcolsep}{0.5mm}
\centering
    \begin{tabular}{ l | c c }
    \hline
    & $(topic_{self}, topic_{oppo}) \rightarrow (topic_{self}^{\prime}, topic_{oppo}^{\prime})$ & \textit{Percent}\\ 
    
    \hline
	\multirow{3}{3.5em}{Winners} & (Strong, Weak) $\rightarrow$ (Strong, Weak) & 12.7\% \\
	& (Strong, Strong) $\rightarrow$ (Strong, Strong) & 10.5\% \\ 
	& (Strong, Strong) $\rightarrow$ (Strong, Weak) & 8.9\% \\					

	\hline
	\hline
	
	\multirow{3}{3.5em}{Losers} & (Weak, Strong) $\rightarrow$ (Weak, Strong) & 11.9\% \\
	& (Strong, Strong) $\rightarrow$ (Strong, Strong) & 9.0\% \\
	& (Strong, Weak) $\rightarrow$ (Strong, Weak) & 8.8\% \\

	\hline
	\end{tabular}
	}
	\vspace{-3mm}
    \caption{\small Top 3 types of shifts. $topic_{self}$ and $topic_{oppo}$ are the strengths of the current topic for one side and their opponent. $topic_{self}^{\prime}$ and $topic_{oppo}^{\prime}$ are the strengths of the topic for the following arguments. }
	\label{tab:topic_shift_finegrain}
\end{table}

A more sophisticated strategy is if the debaters also attempt to put their opponents at a disadvantage with topic shifts. We consider the topic strengths of a current argument for both the speaker (``\textit{self}") and their ``\textit{opponent}", as well as the strength of the following argument.
The top 3 types of shifts are listed in Table~\ref{tab:topic_shift_finegrain}. As can be seen, winners are more likely to be in a strong (for them) and weak (for the opponent) situation and to stay there, while losers are more likely to be in the reverse.  Both sides generally stay in the same strength configuration from argument 1 to argument 2, but winners are also likely (row 3) to employ the strategy of shifting from a topic that is strong for both sides, to one that is strong for them and weak for the opponent.

%\begin{table*}[t]
%    {\fontsize{8}{9}\selectfont
%    \setlength{\baselineskip}{0pt}
%\centering
%    \begin{tabular}{| l | l l | l l |}
%    \hline
%    	& \multicolumn{2}{c|}{\textbf{Winners}} & \multicolumn{2}{c|}{\textbf{Losers}}\\
%     & $(topic_{t}^{self}, topic_{t}^{oppo}) \rightarrow (topic_{t+1}^{self}, topic_{t+1}^{oppo})$ & \textit{Percent} & $(topic_{t}^{self}, topic_{t}^{oppo}) \rightarrow (topic_{t+1}^{self}, topic_{t+1}^{oppo})$ & \textit{Percent}\\
%     \hline
%\multirow{3}{4em}{Top 3 Shift} & (Strong, Weak) $\rightarrow$ (Strong, Weak) & 12.7\% & (Weak, Strong) $\rightarrow$ (Weak, Strong) & 11.9\%\\
%     & (Strong, Strong) $\rightarrow$ (Strong, Strong) & 10.5\% 						& (Strong, Strong) $\rightarrow$ (Strong, Strong) & 9.0\%\\
%     & (Strong, Strong) $\rightarrow$ (Strong, Weak) & 8.9\% 							& (Strong, Weak) $\rightarrow$ (Strong, Weak) & 8.8\%\\
%     \hline
%     \hline
%\multirow{3}{4em}{Bottom 3 Shift} & (Weak, Weak) $\rightarrow$ (Weak, Weak) & 3.2\% & (Weak, Weak) $\rightarrow$ (Weak, Weak) & 2.9\% \\
%     & (Weak, Strong) $\rightarrow$ (Weak, Weak) & 3.6\% & (Weak, Weak) $\rightarrow$ (Strong, Weak) & 3.8\% \\
%     & (Weak, Weak) $\rightarrow$ (Weak, Strong) & 3.8\% & (Strong, Weak) $\rightarrow$ (Weak, Weak) & 4.0\% \\
%
%	\hline
%	\end{tabular}
%	}
%	\vspace{-2mm}
%    \caption{\small Topic shifting behavior.}
%	\label{tab:topic_shift_finegrain}
%\end{table*}

\subsection{Feature Analysis}
\label{sec:feature_analysis}

\begin{table}[t]
\hspace{-3mm}
    {\fontsize{8}{9}\selectfont
    \setlength{\baselineskip}{0pt}
    \setlength{\tabcolsep}{0.0mm}    
%    \begin{tabular}{| l | p{30mm} p{30mm} |}
     \begin{tabular}{ l | l l }
    \hline
 	\multirow{2}{6em}{\textbf{Category}} & \multicolumn{2}{c}{\textbf{Topic Strength}}\\ 
 	
 	 & \textsc{Strong} & \textsc{Weak} \\
 	 \hline
 	\multirow{7}{6em}{\textsc{Basic}} &	\# ``we"$_{full}$	&	\# ``you"$_{inter}$ $\ast$	\\
 	&	\# ``they"$_{inter}$	&	\# ``I"$_{inter}$	\\
 	&	\# ``emotion:sadness"$_{full}$	&	\# ``emotion:joy"$_{full}$	$\ast$\\
 	&	\# ``emotion:disgust"$_{full}$	&	\# ``emotion:trust"$_{full}$ $\ast$ $\ast$\\	
% 	&	\# unique adverbs$_{full}$ $\ast$	&	\# unique adverbs$_{full}$ $\ast$	\\	
% 	&	\# verbs$_{inter}$	&	\# verbs$_{inter}$	\\
 	&		&	\# unique nouns$_{full}$ $\ast$	\\
 	&		&	\# ``numbers"$_{full}$	\\		 	
 	&		&	\# named entities$_{inter}$ $\ast$\\	 	
 	\hline
 	
 	\multirow{5}{6em}{\textsc{Style, Semantic, Discourse}}&	\# non-verb hedging	$_{full}$ &	\# non-verb hedging	$_{full}$	\\	
 	&	avg concreteness$_{full}$ $\ast$	& avg arousal score$_{full}$ $\ast$		\\	 	
 	&	\# formal words$_{full}$ $\ast$	&	\# PDTB:temporal$_{inter}$ $\ast$	\\	 	
 	&	\# FS:capability$_{full}$ 	&	\# PDTB:contrast$_{inter}$	\\	
 	&	\# FS:information$_{full}$	&	\# FS:certainty$_{full}$	\\		
	\hline 	
 	
 	\multirow{5}{6em}{\textsc{Sentence, Argument}} &	Flesch Reading Ease$_{full}$ 	&	Flesch Reading Ease$_{full}$	\\	
 	&	\# sentiment:negative$_{full}$ $\ast$	&	\# sentiment:neutral$_{inter}$ $\ast$	\\	
 	&	\# question$_{full}$ &	\# question$_{full}$	\\	 	
 	&	\# audience laughter$_{inter}$ $\ast$	&		\\	 	 	 	 	 	 	 	
 	&	decayed argument count$_{full}$ $\ast$ &		\\	
  	
  	\hline
  	
  	\multirow{2}{6em}{\textsc{Interaction}} &	\# words addressing 	&	if addressing 	\\
  	&	~~~~opponent's argument$_{full}$	&	~~~~opponent's argument$_{full}$\\
  	&	\# common words with 	&	\\
  	&	~~~~opponent's argument$_{full}$ & 	\\
  	 	
	\hline
	\end{tabular}
	}
	\vspace{-3mm}
    \caption{\small Top weighted features joint with topic strength. ``$full$" and ``$inter$" indicates features that are calculated for full debates or the interactive (discussion) phase only. ``FS" denotes frame semantic. Two-sided Mann-Whitney rank test is conducted on between features of wining and losing sides ($\ast$: $p<0.05$, $\ast$ $\ast$: $p<0.01$).}
	\label{tab:features_topweights}
\end{table}

%Lastly, we investigate \textit{which salient features affect the audience} and \textit{how their usage differs across strong and weak topics}. %Analysis in this section is based on our model (topic strength initialized by frequency) trained with the full debate dataset. 

Lastly, we investigate the \textit{linguistic features associated with topics of different strengths that affect the audience}. 
Table~\ref{tab:features_topweights} displays some of the 50 highest weighted features that interact with strong and weak topics. 
Personal pronoun usage has been found to be related to communicative goals in many previous studies~\cite{brown1960pronouns,wilson1990politically}. We find that strong topics are associated with more first person plurals, potentially an indicator of group responsibility~\cite{wilson1990politically}. On the other hand, our model finds that weak topics are associated with second person pronouns, which may be arguments either attacking other discussants or addressing the audience~\cite{simons2011persuasion}. 
For sentiment, previous work~\cite{tan+etal:16a} has found that persuasive arguments are more negative in online discussions. Our model associates negative sentiment and anger words with strong topics, and neutral and joyful languages with weak topics.

\begin{figure}[ht]
\hspace{-3mm}
\stackunder[0pt]{\includegraphics[width=42mm,height=34mm]{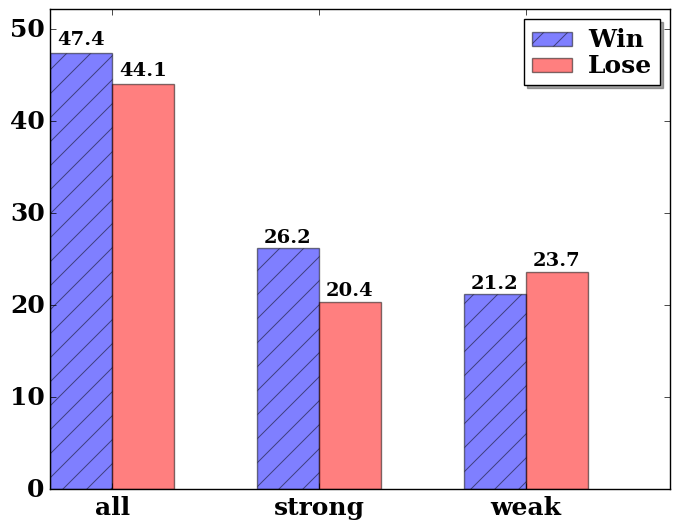}}{\fontsize{8}{8}\selectfont (a) Usage of ``we"}%
\stackunder[0pt]{\includegraphics[width=42mm,height=34mm]{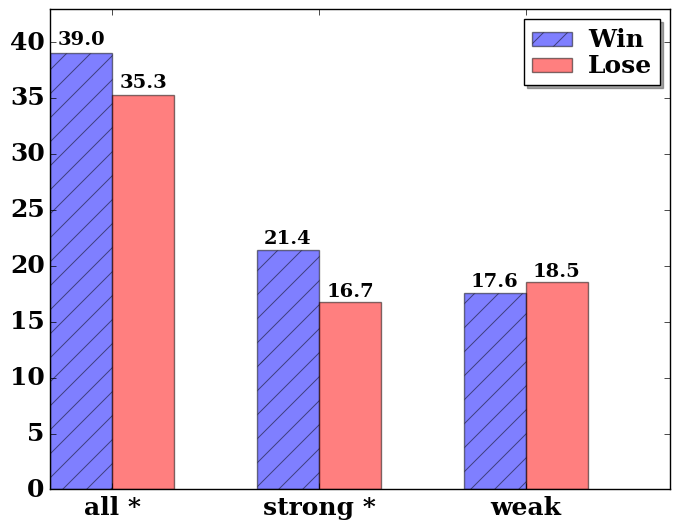}}{\fontsize{8}{8}\selectfont (b) Usage of formal words}

\hspace{-3mm}
\stackunder[0pt]{\includegraphics[width=42mm,height=34mm]{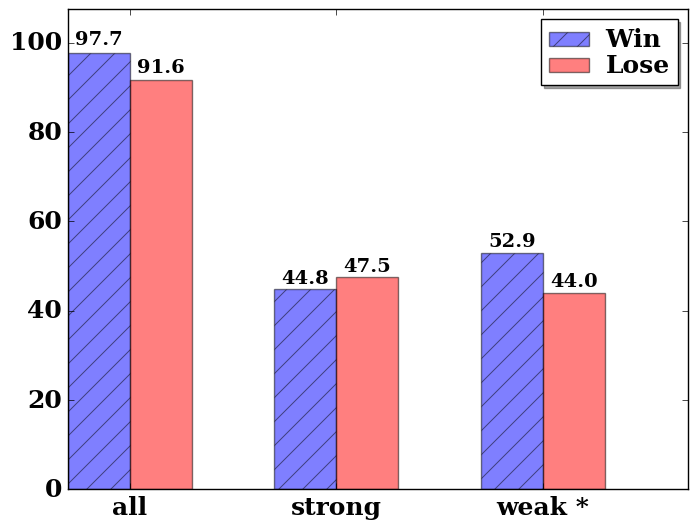}}{\fontsize{8}{8}\selectfont (c) Usage of numbers}%
\stackunder[0pt]{\includegraphics[width=42mm,height=34mm]{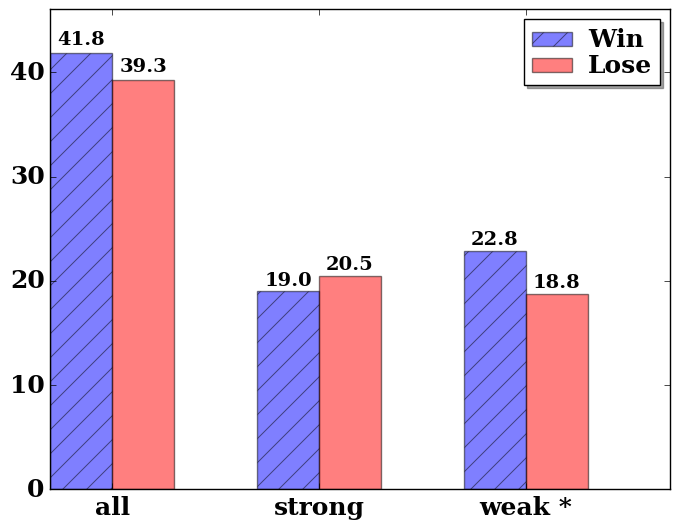}}{\fontsize{8}{8}\selectfont (d) Usage of contrast discourse}

\vspace{-3mm}
\caption{\small Values of sample features with substantial difference between weights associated with ``strong" and ``weak" topics are plotted next to feature values of ``all" arguments. Two-sided Mann-Whitney rank test is conducted between wining and losing sides ($\ast$: $p<0.05$). }
\label{fig:features_diffweights}
\end{figure}

In terms of style and discourse, debaters tend to use more formal and more concrete words for arguments with strong topics. By contrast, arguments with weak topics show more frequent usage of words with intense emotion (higher arousal scores), and contrast discourse connectives. Figure~\ref{fig:features_diffweights} shows how some of these features differ between winners and losers, illustrating the effects on outcome via strong or weak arguments in particular.

Interaction features also play an important role for affecting audiences' opinions. In particular, debaters spend more time (i.e. use more words) addressing their opponents' arguments if it is a strong topic for their opponents. But even for weak topics, it appears helpful to address opponents' arguments.

%Finally, Figure~\ref{fig:features_diffweights} shows how some of these feature weights differ between winning and losing sides, illustrating how the effects on outcome work via strong or weak arguments in particular.

%Without considering topic strength, interactions between debaters are not found among the top 50 weighted features. But 
%For topic strength-specific features, 
%When a topic is strong for both sides, we find the interaction feature that measures the number of words used to address an opponent's argument is very salient (among the top 10). 
%
%Finally, we show the incremental improvement gained by adding in different subcategories of features as follows. %Note the jumps in accuracy obtained by adding in topic strength (``Sentence-level" and ``Argument-level") and then again from adding in the content-rhetoric interactions.

%% file: related.tex
% para1: argumentation mining in general, specifically argument extraction and classification. however, very little of existing work tries to cluster arguments into topics according to their standpoint (is there any?)
%Building off the structural theories of \newcite{toulmin1958uses}, \newcite{cohen1987}, and \newcite{walton2008argumentation} argument mining has successfully developed to recognize, reconstruct, and classify arguments in text. 

Previous work on debate and persuasion has studied the dynamic of audience response to debates and the rhetorical frames the speakers use~\cite{boydstun2014realtime}. However, this work is limited by the scarcity of data and does not focus on the interactions between content and language usage. %Recently, there has been a growing interest in understanding the relation between arguments' content and their persuasion effect~\cite{tan+etal:16a,canobasave-he:2016:N16-1}. 
Topic control, operationalized as the tendency of one side to adopt or avoid the words preferentially used by the other, is investigated in \newcite{zhangetal:NAACL2016} to predict debate outcome using the Intelligence Squared data. %They test whether using the other side's ``talking points'' carries predictive power of debate outcome. 
Our work complements theirs in examining topic interactions, but brings an additional focus on the latent persuasive strength of topics, as well as strength interactions. 
\newcite{tan+etal:16a} examine various structural and linguistic features associated with persuasion on Reddit; they find that some topics correlates more with malleable opinions. Here we develop a more general model of latent topic strength and the linguistic features associated with strength. 
%They ``de-emphasize \textit{what} is being said, in favor of \textit{how} it is expressed,'' but conclude that ``better semantic models...could open up deeper investigations into effective persuasion strategies.''  Our work attempts to incorporate some of these semantic effects. 
% 
%Topic control has also been studied for political debates. For instance, topic shift behavior is modeled by a nonparametric hierarchical Bayesian model in \newcite{nguyen2014modeling}. \newcite{prabhakaran-arora-rambow:2014:EMNLP2014} find that changing topics in presidential primary debates positively correlates with the candidates' power, which is measured based on their relative standing in recent public polls. This work complements previous research in examining effects of topic interactions on debate success. 

%We complement their work with argument exchange understanding on topic level, which avoids ambiguity of word usage. %\newcite{habernal-gurevych:2016:P16-1}  study the relative persuasiveness of pairwise arguments on the same topic by using sentence embeddings output by recurrent neural networks. 
%However, none of them considers the inherent topical strength of arguments, nor its interplay with style and the opposing side.

Additional work has focused on the influence of agenda setting --- controlling which topics are discussed~\cite{nguyen2014modeling}, and framing~\cite{card2015media,tsur2015frame} --- emphasizing certain aspects or interpretations of an issue. \newcite{greene2009syntactic} study the syntactic aspects of framing, where syntactic choices are found to correlate with the sentiment perceived by readers. 
%Using a corpus of congressional texts, \newcite{tsur2015frame} discover complex interactions between the processes of framing, attention shifts and agenda setting. 
%\newcite{nguyen2014modeling} explore the power dynamics of topic control, and \newcite{prabhakaran2014power} and \newcite{prabhakaran2014topic} find that topic shifting tendencies are correlated with a speaker's power. 
%Topic shift behavior is modeled by a nonparametric hierarchical Bayesian model in \newcite{nguyen2014modeling}. 
Based on the topic shifting model of \newcite{nguyen2014modeling}, \newcite{prabhakaran-arora-rambow:2014:EMNLP2014} find that changing topics in presidential primary debates positively correlates with the candidates' power, which is measured based on their relative standing in recent public polls. 
This supports our finding that both sides \textit{seek} to shift topics, but that winners are more likely to shift and shift to topics which are strong for them but weak for their opponents.

Our work is in line with argumentation mining. Existing work in this area focuses on argument extraction~\cite{moens2007automatic,palau2009argumentation,mochales2011argumentation} and argument scheme classification~\cite{biran2011identifying,feng2011classifying,rooney2012applying,stab2014identifying}. Though stance prediction has also been studied~\cite{thomas2006get,hasan2014you}, we are not aware of any work that extracts arguments according to topics and position. 
Argument strength prediction is also studied largely in the domain of student essays \cite{higgins2004evaluating,stab2014identifying,persin2015modeling}. %A strength prediction model, as proposed in~\newcite{persin2015modeling}, primarily evaluates a statement's coherence in relationship to a core argument. 
Notably, none of these distinguishes an argument's strength from its linguistic surface features. This is a gap we aim to fill.

%% file: conclusion.tex
We present a debate prediction model that learns latent persuasive strengths of topics, linguistic style of arguments, and the interactions between the two. Experiments on debate outcome prediction indicate that our model outperforms comparisons using audience responses or linguistic features alone. Our model also shows that winners use stronger arguments and strategically shift topics to stronger ground. We also find that strong and weak arguments differ in their language usage in ways relevant to various behavioral theories of persuasion. 
%Perhaps reassuringly for democracy and deliberation, better arguments do seem to win debates.  

%\paragraph{Future Work.}
% continuous variable

% variation across speakers

% opinion change and topic strength distribution, whether people get more impressed and thus change opinion when one side has less strong topics

% type of arguments: evidence, logical reasoning

%% file: main.bbl
\begin{thebibliography}{}

\bibitem[\protect\citename{Baumgartner \bgroup et al.\egroup
  }2008]{baumgartner2008decline}
Frank~R. Baumgartner, Suzanna~L. De~Boef, and Amber~E. Boydstun.
\newblock 2008.
\newblock {\em The decline of the death penalty and the discovery of
  innocence}.
\newblock Cambridge University Press.

\bibitem[\protect\citename{Bender \bgroup et al.\egroup
  }2011]{bender2011annotating}
Emily~M. Bender, Jonathan~T. Morgan, Meghan Oxley, Mark Zachry, Brian
  Hutchinson, Alex Marin, Bin Zhang, and Mari Ostendorf.
\newblock 2011.
\newblock Annotating social acts: Authority claims and alignment moves in
  wikipedia talk pages.
\newblock In {\em Proceedings of the Workshop on Language in Social Media (LSM
  2011)}, pages 48--57, Portland, Oregon, June. Association for Computational
  Linguistics.

\bibitem[\protect\citename{Biran and Rambow}2011]{biran2011identifying}
Or~Biran and Owen Rambow.
\newblock 2011.
\newblock Identifying justifications in written dialogs by classifying text as
  argumentative.
\newblock {\em International Journal of Semantic Computing}, 5(04):363--381.

\bibitem[\protect\citename{Boydstun \bgroup et al.\egroup
  }2014]{boydstun2014realtime}
Amber~E. Boydstun, Rebecca~A. Glazier, Matthew~T. Pietryka, and Philip Resnik.
\newblock 2014.
\newblock Real-time reactions to a 2012 presidential debate a method for
  understanding which messages matter.
\newblock {\em Public Opinion Quarterly}, 78(S1):330--343.

\bibitem[\protect\citename{Brooke \bgroup et al.\egroup
  }2010]{Brooke:2010:AAL:1944566.1944577}
Julian Brooke, Tong Wang, and Graeme Hirst.
\newblock 2010.
\newblock Automatic acquisition of lexical formality.
\newblock In {\em Proceedings of the 23rd International Conference on
  Computational Linguistics: Posters}, COLING '10, pages 90--98, Stroudsburg,
  PA, USA. Association for Computational Linguistics.

\bibitem[\protect\citename{Brown and Gilman}1960]{brown1960pronouns}
Roger Brown and Albert Gilman, 1960.
\newblock {\em The pronouns of power and solidarity}, pages 253--276.
\newblock MIT Press, Cambridge, MA.

\bibitem[\protect\citename{Brysbaert \bgroup et al.\egroup
  }2014]{brysbaert2014concreteness}
Marc Brysbaert, Amy~Beth Warriner, and Victor Kuperman.
\newblock 2014.
\newblock Concreteness ratings for 40 thousand generally known english word
  lemmas.
\newblock {\em Behavior research methods}, 46(3):904--911.

\bibitem[\protect\citename{Cano-Basave and He}2016]{canobasave-he:2016:N16-1}
Amparo~Elizabeth Cano-Basave and Yulan He.
\newblock 2016.
\newblock A study of the impact of persuasive argumentation in political
  debates.
\newblock In {\em Proceedings of the 2016 Conference of the North American
  Chapter of the Association for Computational Linguistics: Human Language
  Technologies}, pages 1405--1413, San Diego, California, June. Association for
  Computational Linguistics.

\bibitem[\protect\citename{Card \bgroup et al.\egroup }2015]{card2015media}
Dallas Card, Amber~E. Boydstun, Justin~H. Gross, Philip Resnik, and Noah~A.
  Smith.
\newblock 2015.
\newblock The media frames corpus: Annotations of frames across issues.
\newblock In {\em Proceedings of the 53rd Annual Meeting of the Association for
  Computational Linguistics and the 7th International Joint Conference on
  Natural Language Processing (Volume 2: Short Papers)}, pages 438--444,
  Beijing, China, July. Association for Computational Linguistics.

\bibitem[\protect\citename{Chang \bgroup et al.\egroup
  }2010]{Chang:2010:DLO:1857999.1858065}
Ming-Wei Chang, Dan Goldwasser, Dan Roth, and Vivek Srikumar.
\newblock 2010.
\newblock Discriminative learning over constrained latent representations.
\newblock In {\em Human Language Technologies: The 2010 Annual Conference of
  the North American Chapter of the Association for Computational Linguistics},
  pages 429--437, Los Angeles, California, June. Association for Computational
  Linguistics.

\bibitem[\protect\citename{Cohen}1989]{cohen1989deliberation}
Joshua Cohen.
\newblock 1989.
\newblock {\em Deliberation and Democratic Legitimacy}.
\newblock The Good Polity: Normative Analysis of the State. Basil Blackwell.

\bibitem[\protect\citename{Das \bgroup et al.\egroup }2014]{das2014frame}
Dipanjan Das, Desai Chen, Andr{\'e}~F.T. Martins, Nathan Schneider, and Noah~A.
  Smith.
\newblock 2014.
\newblock Frame-semantic parsing.
\newblock {\em Computational Linguistics}, 40(1):9--56.

\bibitem[\protect\citename{Dryzek and List}2003]{dryzek2003social}
John~S. Dryzek and Christian List.
\newblock 2003.
\newblock Social choice theory and deliberative democracy: A reconciliation.
\newblock {\em British Journal of Political Science}, 33(01):1 -- 28.

\bibitem[\protect\citename{Durik \bgroup et al.\egroup }2008]{durik2008effects}
Amanda~M. Durik, M.~Anne Britt, Rebecca Reynolds, and Jennifer Storey.
\newblock 2008.
\newblock The effects of hedges in persuasive arguments: A nuanced analysis of
  language.
\newblock {\em Journal of Language and Social Psychology}.

\bibitem[\protect\citename{Feng and Hirst}2011]{feng2011classifying}
Vanessa~Wei Feng and Graeme Hirst.
\newblock 2011.
\newblock Classifying arguments by scheme.
\newblock In {\em Proceedings of the 49th Annual Meeting of the Association for
  Computational Linguistics: Human Language Technologies}, pages 987--996,
  Portland, Oregon, USA, June. Association for Computational Linguistics.

\bibitem[\protect\citename{Fillmore}1976]{fillmore1976frame}
Charles~J. Fillmore.
\newblock 1976.
\newblock Frame semantics and the nature of language.
\newblock {\em Annals of the New York Academy of Sciences}, 280(1):20--32.

\bibitem[\protect\citename{Goldwasser and
  Daum\'{e}~III}2014]{goldwasser2014object}
Dan Goldwasser and Hal Daum\'{e}~III.
\newblock 2014.
\newblock ``{I} object!" modeling latent pragmatic effects in courtroom
  dialogues.
\newblock In {\em Proceedings of the 14th Conference of the European Chapter of
  the Association for Computational Linguistics}, pages 655--663, Gothenburg,
  Sweden, April. Association for Computational Linguistics.

\bibitem[\protect\citename{Greene and Resnik}2009]{greene2009syntactic}
Stephan Greene and Philip Resnik.
\newblock 2009.
\newblock More than words: Syntactic packaging and implicit sentiment.
\newblock In {\em Proceedings of Human Language Technologies: The 2009 Annual
  Conference of the North American Chapter of the Association for Computational
  Linguistics}, pages 503--511, Boulder, Colorado, June. Association for
  Computational Linguistics.

\bibitem[\protect\citename{Gruber \bgroup et al.\egroup
  }2007]{gruber2007hidden}
Amit Gruber, Yair Weiss, and Michal Rosen-Zvi.
\newblock 2007.
\newblock Hidden topic markov models.
\newblock In {\em International conference on artificial intelligence and
  statistics}, pages 163--170.

\bibitem[\protect\citename{Habermas}1984]{habermas1984theory}
J{\"u}rgen Habermas.
\newblock 1984.
\newblock {\em The theory of communicative action}.
\newblock Beacon Press, Boston.

\bibitem[\protect\citename{Hasan and Ng}2014]{hasan2014you}
Kazi~Saidul Hasan and Vincent Ng.
\newblock 2014.
\newblock Why are you taking this stance? identifying and classifying reasons
  in ideological debates.
\newblock In {\em Proceedings of the 2014 Conference on Empirical Methods in
  Natural Language Processing (EMNLP)}, pages 751--762, Doha, Qatar, October.
  Association for Computational Linguistics.

\bibitem[\protect\citename{Higgins \bgroup et al.\egroup
  }2004]{higgins2004evaluating}
Derrick Higgins, Jill Burstein, Daniel Marcu, and Claudia Gentile.
\newblock 2004.
\newblock Evaluating multiple aspects of coherence in student essays.
\newblock In {\em Human Language Technology Conference of the North American
  Chapter of the Association for Computational Linguistics, Boston,
  Massachusetts, USA, May 2-7, 2004}, pages 185--192.

\bibitem[\protect\citename{Hyland}2005]{metadiscourse10616}
Ken Hyland.
\newblock 2005.
\newblock {\em {Metadiscourse: Exploring interaction in writing}}.
\newblock Continuum, London.

\bibitem[\protect\citename{Irvine}1979]{irvine1979formality}
Judith~T. Irvine.
\newblock 1979.
\newblock Formality and informality in communicative events.
\newblock {\em American Anthropologist}, 81(4):773--790.

\bibitem[\protect\citename{Katzav and Reed}2008]{katzav2008modelling}
Joel Katzav and Chris Reed.
\newblock 2008.
\newblock Modelling argument recognition and reconstruction.
\newblock {\em Journal of Pragmatics}, 40(1):155--172.

\bibitem[\protect\citename{Klein and
  Manning}2003]{Klein:2003:AUP:1075096.1075150}
Dan Klein and Christopher~D. Manning.
\newblock 2003.
\newblock Accurate unlexicalized parsing.
\newblock In {\em Proceedings of the 41st Annual Meeting of the Association for
  Computational Linguistics}, pages 423--430, Sapporo, Japan, July. Association
  for Computational Linguistics.

\bibitem[\protect\citename{Mansbridge}2003]{mansbridge2003rethinking}
Jane Mansbridge.
\newblock 2003.
\newblock Rethinking representation.
\newblock {\em The American Political Science Review}, 97(04):515--528.

\bibitem[\protect\citename{Mansbridge}2015]{mansbridge2015minimalist}
Jane Mansbridge, 2015.
\newblock {\em A Minimalist Definition of Deliberation}, book section~2, pages
  27--50.
\newblock Equity and Development series. World Bank Publications.

\bibitem[\protect\citename{McCombs}2005]{mccombs2005look}
Maxwell McCombs.
\newblock 2005.
\newblock A look at agenda-setting: Past, present and future.
\newblock {\em Journalism studies}, 6(4):543--557.

\bibitem[\protect\citename{Mochales and Moens}2011]{mochales2011argumentation}
Raquel Mochales and Marie-Francine Moens.
\newblock 2011.
\newblock Argumentation mining.
\newblock {\em Artificial Intelligence and Law}, 19(1):1--22.

\bibitem[\protect\citename{Moens \bgroup et al.\egroup
  }2007]{moens2007automatic}
Marie-Francine Moens, Erik Boiy, Raquel~Mochales Palau, and Chris Reed.
\newblock 2007.
\newblock Automatic detection of arguments in legal texts.
\newblock In {\em Proceedings of the 11th international conference on
  Artificial intelligence and law}, pages 225--230. ACM.

\bibitem[\protect\citename{Mohammad and Turney}2013]{Mohammad13}
Saif~M. Mohammad and Peter~D. Turney.
\newblock 2013.
\newblock Crowdsourcing a word-emotion association lexicon.
\newblock {\em Computational Intelligence}, 29(3):436--465.

\bibitem[\protect\citename{Nguyen \bgroup et al.\egroup
  }2014]{nguyen2014modeling}
Viet-An Nguyen, Jordan Boyd-Graber, Philip Resnik, Deborah~A. Cai, Jennifer~E.
  Midberry, and Yuanxin Wang.
\newblock 2014.
\newblock Modeling topic control to detect influence in conversations using
  nonparametric topic models.
\newblock {\em Machine Learning}, 95(3):381--421.

\bibitem[\protect\citename{Nguyen \bgroup et al.\egroup
  }2015]{nguyen-EtAl:2015:ACL-IJCNLP2}
Viet-An Nguyen, Jordan Boyd-Graber, Philip Resnik, and Kristina Miler.
\newblock 2015.
\newblock Tea party in the house: A hierarchical ideal point topic model and
  its application to republican legislators in the 112th congress.
\newblock In {\em Proceedings of the 53rd Annual Meeting of the Association for
  Computational Linguistics and the 7th International Joint Conference on
  Natural Language Processing (Volume 1: Long Papers)}, pages 1438--1448,
  Beijing, China, July. Association for Computational Linguistics.

\bibitem[\protect\citename{Noelle-Neumann}1974]{noelleNeumann1974}
Elisabeth Noelle-Neumann.
\newblock 1974.
\newblock The spiral of silence a theory of public opinion.
\newblock {\em Journal of communication}, 24(2):43--51.

\bibitem[\protect\citename{Palau and Moens}2009]{palau2009argumentation}
Raquel~Mochales Palau and Marie-Francine Moens.
\newblock 2009.
\newblock Argumentation mining: the detection, classification and structure of
  arguments in text.
\newblock In {\em Proceedings of the 12th international conference on
  artificial intelligence and law}, pages 98--107. ACM.

\bibitem[\protect\citename{Persing and Ng}2015]{persin2015modeling}
Isaac Persing and Vincent Ng.
\newblock 2015.
\newblock Modeling argument strength in student essays.
\newblock In {\em Proceedings of the 53rd Annual Meeting of the Association for
  Computational Linguistics and the 7th International Joint Conference on
  Natural Language Processing (Volume 1: Long Papers)}, pages 543--552,
  Beijing, China, July. Association for Computational Linguistics.

\bibitem[\protect\citename{Prabhakaran \bgroup et al.\egroup
  }2014]{prabhakaran-arora-rambow:2014:EMNLP2014}
Vinodkumar Prabhakaran, Ashima Arora, and Owen Rambow.
\newblock 2014.
\newblock Staying on topic: An indicator of power in political debates.
\newblock In {\em Proceedings of the 2014 Conference on Empirical Methods in
  Natural Language Processing (EMNLP)}, pages 1481--1486, Doha, Qatar, October.
  Association for Computational Linguistics.

\bibitem[\protect\citename{Prasad \bgroup et al.\egroup }2007]{prasad2007penn}
Rashmi Prasad, Eleni Miltsakaki, Nikhil Dinesh, Alan Lee, Aravind Joshi, Livio
  Robaldo, and Bonnie~L. Webber.
\newblock 2007.
\newblock The penn discourse treebank 2.0 annotation manual.

\bibitem[\protect\citename{Rawls}1997]{rawls1997idea}
John Rawls.
\newblock 1997.
\newblock The idea of public reason revisited.
\newblock {\em The University of Chicago Law Review}, 64(3):765--807.

\bibitem[\protect\citename{R\"{o}der \bgroup et al.\egroup
  }2015]{Roder:2015:EST:2684822.2685324}
Michael R\"{o}der, Andreas Both, and Alexander Hinneburg.
\newblock 2015.
\newblock Exploring the space of topic coherence measures.
\newblock In {\em Proceedings of the Eighth ACM International Conference on Web
  Search and Data Mining}, WSDM '15, pages 399--408, New York, NY, USA. ACM.

\bibitem[\protect\citename{Rooney \bgroup et al.\egroup
  }2012]{rooney2012applying}
Niall Rooney, Hui Wang, and Fiona Browne.
\newblock 2012.
\newblock Applying kernel methods to argumentation mining.
\newblock In {\em Proceedings of the Twenty-Fifth International Florida
  Artificial Intelligence Research Society Conference, Marco Island, Florida.
  May 23-25, 2012}.

\bibitem[\protect\citename{Simons and Jones}2011]{simons2011persuasion}
Herbert~W. Simons and Jean Jones.
\newblock 2011.
\newblock {\em Persuasion in society}.
\newblock Routledge, 2nd ed. edition.

\bibitem[\protect\citename{Socher \bgroup et al.\egroup
  }2013]{socher-EtAl:2013:EMNLP}
Richard Socher, Alex Perelygin, Jean~Y. Wu, Jason Chuang, Christopher~D.
  Manning, Andrew~Y. Ng, and Christopher Potts.
\newblock 2013.
\newblock Recursive deep models for semantic compositionality over a sentiment
  treebank.
\newblock In {\em Proceedings of the 2013 Conference on Empirical Methods in
  Natural Language Processing (EMNLP)}, pages 1631--1642, Seattle, Washington,
  USA, October. Association for Computational Linguistics.

\bibitem[\protect\citename{Stab and Gurevych}2014]{stab2014identifying}
Christian Stab and Iryna Gurevych.
\newblock 2014.
\newblock Identifying argumentative discourse structures in persuasive essays.
\newblock In {\em Proceedings of the 2014 Conference on Empirical Methods in
  Natural Language Processing (EMNLP)}, pages 46--56, Doha, Qatar, October.
  Association for Computational Linguistics.

\bibitem[\protect\citename{Sunstein}1999]{sunstein1999}
Cass~R. Sunstein.
\newblock 1999.
\newblock The law of group polarization.

\bibitem[\protect\citename{Tan \bgroup et al.\egroup }2016]{tan+etal:16a}
Chenhao Tan, Vlad Niculae, Cristian Danescu{-}Niculescu{-}Mizil, and Lillian
  Lee.
\newblock 2016.
\newblock Winning arguments: Interaction dynamics and persuasion strategies in
  good-faith online discussions.
\newblock In {\em Proceedings of the 25th International Conference on World
  Wide Web, {WWW} 2016, Montreal, Canada, April 11 - 15, 2016}, pages 613--624.

\bibitem[\protect\citename{Thomas \bgroup et al.\egroup }2006]{thomas2006get}
Matt Thomas, Bo~Pang, and Lillian Lee.
\newblock 2006.
\newblock Get out the vote: Determining support or opposition from
  congressional floor-debate transcripts.
\newblock In {\em Proceedings of the 2006 Conference on Empirical Methods in
  Natural Language Processing (EMNLP)}, pages 327--335. Association for
  Computational Linguistics.

\bibitem[\protect\citename{Tsur \bgroup et al.\egroup }2015]{tsur2015frame}
Oren Tsur, Dan Calacci, and David Lazer.
\newblock 2015.
\newblock A frame of mind: Using statistical models for detection of framing
  and agenda setting campaigns.
\newblock In {\em Proceedings of the 53rd Annual Meeting of the Association for
  Computational Linguistics and the 7th International Joint Conference on
  Natural Language Processing (Volume 1: Long Papers)}, pages 1629--1638,
  Beijing, China, July. Association for Computational Linguistics.

\bibitem[\protect\citename{Wang and Cardie}2014a]{wang-cardie:2014:W14-26}
Lu~Wang and Claire Cardie.
\newblock 2014a.
\newblock Improving agreement and disagreement identification in online
  discussions with a socially-tuned sentiment lexicon.
\newblock In {\em Proceedings of the 5th Workshop on Computational Approaches
  to Subjectivity, Sentiment and Social Media Analysis}, pages 97--106,
  Baltimore, Maryland, June. Association for Computational Linguistics.

\bibitem[\protect\citename{Wang and Cardie}2014b]{wang-cardie:2014:P14-2}
Lu~Wang and Claire Cardie.
\newblock 2014b.
\newblock A piece of my mind: A sentiment analysis approach for online dispute
  detection.
\newblock In {\em Proceedings of the 52nd Annual Meeting of the Association for
  Computational Linguistics (Volume 2: Short Papers)}, pages 693--699,
  Baltimore, Maryland, June. Association for Computational Linguistics.

\bibitem[\protect\citename{Warriner \bgroup et al.\egroup
  }2013]{warriner2013norms}
Amy~Beth Warriner, Victor Kuperman, and Marc Brysbaert.
\newblock 2013.
\newblock Norms of valence, arousal, and dominance for 13,915 english lemmas.
\newblock {\em Behavior research methods}, 45(4):1191--1207.

\bibitem[\protect\citename{Wilson \bgroup et al.\egroup }2005]{Wilson:2005:RCP}
Theresa Wilson, Janyce Wiebe, and Paul Hoffmann.
\newblock 2005.
\newblock Recognizing contextual polarity in phrase-level sentiment analysis.
\newblock In {\em Proceedings of the Conference on Human Language Technology
  and Empirical Methods in Natural Language Processing}, HLT '05, pages
  347--354, Stroudsburg, PA, USA. Association for Computational Linguistics.

\bibitem[\protect\citename{Wilson}1990]{wilson1990politically}
John Wilson.
\newblock 1990.
\newblock {\em Politically speaking: The pragmatic analysis of political
  language}.
\newblock Basil Blackwell.

\bibitem[\protect\citename{Yu and Joachims}2009]{Yu:2009:LSS:1553374.1553523}
Chun-Nam~John Yu and Thorsten Joachims.
\newblock 2009.
\newblock Learning structural svms with latent variables.
\newblock In {\em Proceedings of the 26th Annual International Conference on
  Machine Learning}, ICML '09, pages 1169--1176, New York, NY, USA. ACM.

\bibitem[\protect\citename{Zhang \bgroup et al.\egroup
  }2016]{zhangetal:NAACL2016}
Justine Zhang, Ravi Kumar, Sujith Ravi, and Cristian Danescu-Niculescu-Mizil.
\newblock 2016.
\newblock Conversational flow in oxford-style debates.
\newblock In {\em Proceedings of the 2016 Conference of the North American
  Chapter of the Association for Computational Linguistics: Human Language
  Technologies}, pages 136--141, San Diego, California, June. Association for
  Computational Linguistics.

\end{thebibliography}
